\documentclass[sigconf]{aamas} 

\usepackage{balance}
\usepackage{subcaption}
\usepackage{graphicx}
\usepackage{dsfont}
\usepackage{algpseudocode}

\def\IR{\mathds{R}}

\newcommand{\A}{\mathcal{A}}

\renewcommand{\H}{\mathcal{H}}
\newcommand{\M}{\mathcal{M}}
\newcommand{\T}{\mathcal{T}}
\newcommand{\X}{\mathcal{X}}

\newcommand{\Z}{\mathcal{Z}}

\newcommand{\eps}{\varepsilon}

\renewcommand{\P}{\mathbf{P}}
\renewcommand{\O}{\mathbf{O}}
\renewcommand{\vec}[1]{{\boldsymbol#1}}

\newcommand{\EE}[2][]{\mathbb{E}_{#1}\left[#2\right]} 
\newcommand{\PP}[2][]{\mathbb{P}_{#1}\left[#2\right]} 

\providecommand{\set}[1]{\ensuremath{\left\{#1\right\}}}
\providecommand{\abs}[1]{\ensuremath{\left\lvert#1\right\rvert}}

\newcommand{\bel}{\mathrm{Bel}}
\newcommand{\KL}{\mathrm{KL}}

\DeclareMathOperator{\argmax}{\rm argmax}

\setcopyright{ifaamas}
\acmConference[AAMAS '22]{Proc.\@ of the 21st International Conference
on Autonomous Agents and Multiagent Systems (AAMAS 2022)}{May 9--13, 2022}
{Auckland, New Zealand}{P.~Faliszewski, V.~Mascardi, C.~Pelachaud,
M.E.~Taylor (eds.)}
\copyrightyear{2022}
\acmYear{2022}
\acmDOI{}
\acmPrice{}
\acmISBN{}

\acmSubmissionID{385}

\title{Assisting Unknown Teammates in Unknown Tasks: \\ Ad Hoc Teamwork under Partial Observability}

\author{Jo\~{a}o G. Ribeiro}
\affiliation{
  \institution{INESC-ID and Instituto Superior T\'{e}cnico}
  \city{Lisbon}
  \country{Portugal}}
\email{joao.g.ribeiro@tecnico.ulisboa.pt}

\author{Cassandro Martinho}
\affiliation{
  \institution{Instituto Superior T\'{e}cnico}
  \city{Lisbon}
  \country{Portugal}}
\email{cassandro.martinho@tecnico.ulisboa.pt}

\author{Alberto Sardinha}
\affiliation{
  \institution{INESC-ID and Instituto Superior T\'{e}cnico}
  \city{Lisbon}
  \country{Portugal}}
\email{jose.alberto.sardinha@tecnico.ulisboa.pt}

\author{Francisco S.\ Melo}
\affiliation{
  \institution{INESC-ID and Instituto Superior T\'{e}cnico}
  \city{Lisbon}
  \country{Portugal}}
\email{fmelo@inesc-id.pt}

\begin{abstract}
In this paper, we present a novel Bayesian online prediction algorithm for the problem setting of ad hoc teamwork under partial observability (ATPO), which enables on-the-fly collaboration with unknown teammates performing an unknown task without needing a pre-coordination protocol. Unlike previous works that assume a fully observable state of the environment, ATPO accommodates partial observability, using the agent's observations to identify which task is being performed by the teammates. Our approach assumes neither that the teammate's actions are visible nor an environment reward signal. We evaluate ATPO in three domains---two modified versions of the Pursuit domain with partial observability and the overcooked domain of \citet{carroll19nips}. Our results show that ATPO is effective and robust in identifying the teammate's task from a large library of possible tasks, efficient at solving it in near-optimal time, and scalable in adapting to increasingly larger problem sizes.
\end{abstract}


\newcommand{\BibTeX}{\rm B\kern-.05em{\sc i\kern-.025em b}\kern-.08em\TeX}

\begin{document}

\pagestyle{fancy}
\fancyhead{}

\maketitle 


\section{Introduction}

The problem of ad hoc teamwork was first proposed by \citet{stone10aaai}, and considers an autonomous agent (the ``ad hoc agent'') deployed into an existing group of ``teammates'',  with whom it must engage in teamwork while having no pre-established communication or coordination protocols. Previous works on ad hoc teamwork rely on strong assumptions regarding the interaction between the ``ad hoc agent'' and the ``teammates''. For example,  \citet{barrett17aij,hu1998multiagent,bu2008comprehensive} consider that a reward signal is available to the agents to learn from, casting the problem of ad hoc teamwork as a specialization of multiagent RL. Other works do not consider an RL setting, instead, assuming that the team is performing one among a set of possible tasks, and use the observed teammate behavior to infer the target task \citep{melo16jaamas,fern2007decision}. Closest to our work, \citet{ribeiro2021helping} recently proposed an approach to ad hoc teamwork where the actions of the teammates are not assumed observable; however, the underlying state of the environment is.

Full state observability is a strong assumption within the ad hoc teamwork setting, as it significantly restricts the applicability of such an approach to very narrow settings. Either all tasks share the state space and dynamics (and thus differ only in their goal) or ---if the dynamics are different---such differences greatly facilitate identifying the underlying task.

In this paper, we address ad hoc teamwork with partial observability. In this setting, the ``ad hoc agent'' can only access a limited view of the environment state and must (i) infer what the underlying target task may be; (ii) infer how the teammates are playing it; (iii) plan how to coordinate with the ``teammates''. Moreover, the agent cannot observe the actions of the teammates nor communicate with them.

The setting of ad hoc teamwork with partial observability significantly broadens the applicability of ad hoc teamwork in real-world scenarios. It allows for a much richer set of possible tasks, with widely different dynamics and states that need only to share the perceptual space of the ad hoc agent. It is, therefore, suited to address tasks involving {\em ad hoc robotic agents}, accommodating the natural perceptual limitations of robotic platforms \citep{genter2017jaamas}.

To address partial observability, we build a model that encapsulates the ``perceptual dynamics'' associated with each task and use the history of observations of the agent to estimate which of the set of possible tasks best matches such perceptual dynamics. Our results in several benchmark multiagent problems from the literature illustrate that our approach, ATPO (Ad hoc Teamwork with Partial Observability), can correctly infer the underlying task and coordinate with the teammates.

In summary, our contributions are threefold:
\begin{itemize}
\item We contribute the formalization of ad hoc teamwork under partial observability;
\item We propose ATPO, a novel approach for ad hoc teamwork with partial observability that infers the underlying target task from the agent's history of observations;
\item We illustrate the applicability of our approach in two partially observable variants of pursuit and in {\em Overcooked}, a recent benchmark for multiagent systems \citep{carroll19nips}.
\end{itemize}


\subsection{Related Work}

Since the seminal work of \citet{stone10aaai}, ad hoc teamwork has been under great attention by the multi-agent systems community. However, in a setting where the main challenges consist on assisting unknown teammates in performing unknown tasks, early works assumed the ad hoc agent knew the target task and the teammates' behavior \citep{stone10aamas,stone09amec,agmon12aamas,genter13aamas,chakraborty13aamas} (reducing the problem to one of planning). 

It wasn't until later on that \cite{melo16jaamas} formalized ad hoc teamwork by breaking it down into three sub-problems (task identification, teammate identification and planning) and allowing authors to specify in their work which they were tackling. Following \cite{melo16jaamas}, \cite{barrett17aij} introduced the PLASTIC framework \citep{barrett17aij, barrett15aaai}, where the authors learn predictive models of the teammates by observing their actions and policies for the tasks via trial-and-error (assuming the environment provides a reward signal). Both assumptions (visible teammate actions and available reward signals), however, limit their approach from being deployed in real-world scenarios.

\cite{ribeiro2021helping} later on tackle these limitation by assuming instead that neither the actions nor the reward signals are available, and instead that the ad hoc agent access to a set of possible tasks and teammates. The authors showcase how by only having access to the state of the environment, the ad hoc agent is able to infer the most likely task/teammate combination. Once more, the assumption that the environment is fully observable will seldom be met in practice since ad hoc agents in the real world will often be plagued by issues of partial observability.

Three similar (yet different) lines of work, parallel to ad hoc teamwork, are the framework of assistance \citep{fern2007decision}, the problem of zero-shot coordination \citep{hu2020other} and the IPOMDP framework \citep{gmytrasiewicz2005framework}. The framework of {\em assistance} \citep{fern2007decision} models a scenario where an agent must help a teammate in solving a given sequential task under uncertainty. The goal, however, is for an agent to assist unknown teammates in performing known tasks. Even though the authors do not refer to the problem as ad hoc teamwork, one could argue that the identification of unknown teammates with known tasks falls under the scope of ad hoc teamwork. In their work, however, \cite{fern2007decision} also consider that the teammate's actions to be accessible to the assistant. The problem of {\em zero-shot coordination} \citep{hu2020other}, unlike in ad hoc teamwork, studies how independently trained agents may interact with one another on first-attempt \citep{lupu2021trajectory,treutlein2021new,bullard2021quasi} (therefore falling out of the scope of ad hoc teamwork). Finally, the IPOMDP framework \citep{gmytrasiewicz2005framework}, considers how an agent can augment the state space of a POMDP taking into account all possible, unknown, teammates. Not only does this approach grow exponentially with the number of possible teammates, but, similar to the framework of assistance, it also assumes that all teammates perform the same task (modeled by a single reward function). In their work, however, \citep{gmytrasiewicz2005framework}, the authors do not assume that the teammates' actions are visible to the agent, therefore sharing a common assumption with our work. Unlike \cite{gmytrasiewicz2005framework}, we expand upon \cite{melo16jaamas} and \cite{ribeiro2021helping}, introducing an approach capable of explicitly identifying the teammate behavior and the tasks' goal by describing the set of possible tasks as a partially observable Markov decision problem and using a Bayesian approach to map the history of observations of the ad hoc agent into a belief over the set of possible tasks and possible teammates (growing linearly with the number of possible teammates and tasks instead of exponentially like IPOMDPs).


\section{Background}%
\label{Sec:Background}

In this work, we address ad hoc teamwork under partial observability using a decision-theoretic framework. This section introduces some key concepts and sets up the nomenclature regarding Markov decision problems and related models.

\subsection{Markov decision problems}

A {\em Markov decision problem} \citep{puterman05}, or MDP, is denoted as a tuple $(\X,\A,\set{\P_a,a\in\A},r,\gamma)$, where $\X$ is the state space, $\A$ is the action space, $\P_a$ is a transition probability matrix, where $\P_a(x'\mid x)$ is probability of moving from state $x$ to $x'$ given action $a\in\A$, $r$ is the expected reward function, and $\gamma\in[0,1]$ is a discount factor.  

A {\em policy} $\pi$ maps states to distributions over actions. We write $\pi(a\mid x)$ to denote the probability of selecting action $a$ in state $x$ according to policy $\pi$. Solving an MDP consists of determining a policy $\pi$ to maximize the value
\begin{equation}
v^\pi(x)\triangleq\EE[A_t\sim\pi(X_t)]{\sum_{t=0}^\infty\gamma^tR_t\mid X_0=x},
\end{equation}
for any initial state $x\in\X$. In the above expression, $X_t$, $A_t$ and $R_t$ denote the (random) state, action and reward at time step $t$. The function $v^\pi:\X\to\IR$ is called a {\em value function}, and a policy $\pi^*$ is {\em optimal} if, given any policy $\pi$, $v^{\pi^*}(x)\geq v^\pi(x)$, for all $x\in\X$. The value function associated with an optimal policy is denoted as $v^*$ and can be computed using, for example, dynamic programming. An optimal policy, $\pi^*$, is such that $\pi^*(a\mid x)>0$ only if $a\in\argmax q^*(x,\cdot)$, where
\begin{equation*}
q^*(x,a)=r(x,a)+\gamma\sum_{x'\in\X}\P_a(x'\mid x)v^*(x').
\end{equation*}


\subsection{Multiagent Markov decision problems}

A {\em multiagent MDP} (MMDP) is an extension of MDPs to multiagent settings and can be described as a tuple 
\begin{equation*}
\M=(N,\X,\set{\A^n,n=1,\ldots,N},\set{\P_{\vec{a}},\vec{a}\in\A},r,\gamma),	
\end{equation*}
where $N$ is the number of agents, $\X$ is the state space, $\A^n$ is the individual action space for agent $n$, $\P_{\vec{a}}$ is the transition probability matrix associated with joint action $\vec{a}$, $r$ is the expected reward function, and $\gamma$ is the discount. We write $\A$ to denote the set of all joint actions, corresponding to the Cartesian product of all individual action spaces $\A^n$, i.e., $\A=\A^1\times\A^2\times\ldots\times\A^N$. We also denote an element of $\A^n$ as $a^n$ and an element of $\A$ as a tuple $\vec{a}=(a^1,\ldots,a^N)$, with $a^n\in\A^n$. We write $\vec{a}^{-n}$ to denote a reduced joint action, i.e., a tuple $\vec{a}^{-n}=(a^1,\ldots,a^{n-1},a^{n+1},\ldots,a^N)$, and thus $\A^{-n}$ is the set of all reduced joint actions. We adopt, for policies, a similar notation. Specifically, we write $\pi^n$ to denote an individual policy for agent $n$, $\vec{\pi}=(\pi^1,\ldots,\pi^N)$ to denote a joint policy, and $\vec{\pi}^{-n}$ to denote a reduced joint policy.

The common goal of the agents in an MMDP is to select a joint policy, $\vec{\pi}^*$, such that $v^{\vec{\pi}^*}(x)\geq v^{\vec{\pi}}(x)$, where, as before, 
\begin{equation}
v^{\vec{\pi}}(x)=\EE[\vec{A}_t\sim\vec{\pi}(X_t)]{\sum_{t=0}^\infty\gamma^tR_t\mid X_0=x}.
\end{equation}
In other words, an MMDP is just an MDP in which the action selection process is distributed across $N$ agents, and can be solved by computing $\vec{\pi}^*$ from $v^*$ as standard MDPs.


\subsection{Partially observable MDPs}

A {\em partially observable MDP}, or POMDP, is an extension of MDPs to partially observable settings. A POMDP can be described as a tuple $(\X,\A,\Z,\set{\P_a,a\in\A},\set{\O_a,a\in\A},r,\gamma)$, where $\X$, $\A$, $\set{\P_a,a\in\A}$, $r$, and $\gamma$, are the same as in MDPs, $\Z$ is the {\em observation space}, and $\O_a$ is the observation probability matrix, where
\begin{equation*}
\O_a(z\mid x)=\PP{Z_{t+1}=z\mid X_{t+1}=x,A_t=a}.	
\end{equation*}
The {\em belief} at time step $t$ is a distribution $\vec{b}_t$ such that
\begin{equation*}
b_t(x)\triangleq\PP{X_t=x\mid X_0\sim\vec{b}_0,A_0=a_0,Z_1=z_1,\ldots,Z_t=z_t},
\end{equation*}
where $\vec{b}_0$ is the initial state distribution. Given the action $a_t$ and the observation $z_{t+1}$, we can update the belief $b_t$ to incorporate the new information yielding
\begin{equation}\label{Eq:Belief-update}
    \begin{split}
  b_{t+1}(x') 
    & =\bel(b_t,a_t,z_{t+1}) \\
    &\triangleq\frac{1}{\rho_{t+1}}\sum_{x\in\X}b_t(x)\P(x'\mid x,a_t)\O(z_{t+1}\mid x',a_t),
\end{split}
\end{equation}
where $\rho_{t+1}$ is a normalization factor. Every finite POMDP admits an equivalent {\em belief-MDP} with $b_t$ being the state of this new MDP at time step $t$. A policy in a POMDP can thus be seen as mapping $\pi$ from beliefs to distributions over actions, and we define
\begin{equation}
v^\pi(b)\triangleq\EE[A_t\sim\pi(b_t)]{\sum_{t=0}^\infty\gamma^tR_t\mid b_0=b}.
\end{equation}
As in MDPs, the value function associated with an optimal policy is denoted as $v^*$ and can be computed using, for example, point-based approaches \citep{pineau06jair}. From $v^*$, the optimal $Q$-function can now be computed as
\begin{multline*}
q^*(b,a)=\sum_{x\in\X}b(x)\Bigg[r(x,a)\\\
+\gamma\sum_{z\in\Z}\sum_{y\in\X}\P(y\mid x,a)\O(z\mid y,a)v^*(\bel(b,a,z))\Bigg],
\end{multline*}
yielding as optimal any policy $\pi^*$ such that $\pi^*(a\mid b)>0$ only if $a\in\argmax_{a\in\A}q^*(b,a)$.

\section{Ad Hoc Teamwork under Partial Observability}

In this section, we introduce our key contributions to ad hoc teamwork under partial observability.


\subsection{Problem formulation}

We consider a team of $N$ agents engaged in a cooperative task (henceforth referred as ``target task''), described as an MMDP $m=(N,\X,\set{\A_n},\set{\P_a},r,\gamma)$. One of the agents (the focus of our work) does not know the task beforehand but must, nevertheless, engage in ad hoc teamwork with the remaining agents to complete the unknown task. We refer to such agent as the ``ad hoc agent'' and denote it as $\alpha$, and refer to the remaining $N-1$ agents collectively as the ``teammates''. Formally, we treat the teammates as a ``meta-agent'' and denote it as $-\alpha$. 

We assume that the teammates all know the target task. The ad hoc agent, however, does not. Instead, it knows only that the target task is one among $K$ possible tasks, where each task can be represented as an MMDP
\begin{equation}
m_k=(2,\X_k,\set{\A^{\alpha},\A_k^{-\alpha}},\set{\P_{k,\vec{a}},\vec{a}\in\A},r_k,\gamma_k).
\end{equation}
Note that we require the action space of the ad hoc agent, $\A^\alpha$, to be the same in all tasks. Other than that, we impose no restrictions on the state space, dynamics, or reward describing these tasks (in particular, they may all be different).

Let $\pi^{-\alpha}_k$ denote a teammates optimal policy for task $k, k=1,\ldots,K$. Then, for task $k$, we have
\begin{multline}\label{Eq:Transitions}
\P_k(y\mid x,a^\alpha)
\triangleq\mathbb{P}[X_{t+1}=y\mid X_t=x,A^\alpha=a^\alpha,\\
A^{-\alpha}\sim\pi^{-\alpha}_k(x),M=m_k],
\end{multline}
for $x,y\in\X_k$, where we write $M=m_k$ to denote the fact that the transitions in \eqref{Eq:Transitions} concerns task $k$.

Let us now suppose that, at each moment, the ad hoc agent cannot observe the underlying state of the environment. Instead, at each time step $t$, the agent can only access an observation $Z_t$. We assume that the observations $Z_t$ take values in a (task-independent) set $\Z$ and depend both on the underlying state of the environment and the previous action of the agent (not the teammates). Specifically, for each task $k=1,\ldots,K$, we assume that there is a family of task-dependent observation probability matrices, $\O_{k,a^\alpha},k=1,\ldots,K,a^\alpha\in\A^\alpha$, with
\begin{equation}
\O_k(z\mid x,a^\alpha)=\PP{Z_t=z\mid X_t=x,A^{\alpha}_{t-1}=a^\alpha,M=m_k}.
\end{equation}
The elements $[\O_{k,a}]_{xz}$ are only defined for $x\in\X_k$. Thus, from the ad hoc agent's perspective, each task $k$ defines a POMDP $\hat{m}_k$ corresponding to the tuple
\begin{equation*}
(\X_k,\A^\alpha,\Z,\set{\P_{k,a^\alpha},a^\alpha\in\A^\alpha},\set{\O_{k,a^\alpha},a^\alpha\in\A^\alpha},r_k,\gamma_k).
\end{equation*}
We denote the solution to $\hat{m}_k$ as $\hat{\pi}_k$.


\subsection{Algorithm}

We adopt a Bayesian framework and treat the target task as a random variable, $M$, taking values in the set of possible MMDP task descriptions, $\M=\set{m_1,\ldots,m_K}$. For $m_k\in\M$, let $p_0(m_k)$ denote the ad hoc agent's prior over $\M$. Additionally, let $H_t$ denote the random variable corresponding to the history of the agent up to time step $t$, defined as
\begin{equation}
H_t=\set{a^\alpha_0,z_1,a^\alpha_1,z_2,\ldots,a^\alpha_{t-1},z_t}.	
\end{equation}
Then, given a history $h_t$, we define 
\begin{equation}
p_t(m_k)\triangleq\PP{M=m_k\mid H_t=h_t}, \qquad m_k\in\M.
\end{equation}
The distribution $p_t$ corresponds to the agent's belief about the target task at time step $t$. The action for the ad hoc agent at time step $t$ can be computed within our Bayesian setting as
\begin{equation}
\nonumber
\pi_t(a^\alpha\mid h_t)
  \triangleq\PP{A^\alpha_t=a^\alpha\mid H_t=h_t}
  =\sum_{k=1}^K\hat{\pi}_k(a^\alpha\mid b_{k,t})p_t(m_k),
\label{Eq:Policy}
\end{equation}
where
\begin{equation}
b_{k,t}(x)\triangleq\PP{X_t=x\mid H_t=h_t,M=m_k},
\end{equation}
for $x\in\X_k$. Upon selecting an action $a^\alpha_t$ and making a new observation $z_{t+1}$, we can update $p_t$ by noting that
\begin{align*}
\lefteqn{p_{t+1}(m_k)}\\
&=\PP{M=m_k\mid H_{t+1}=\set{h_t,a^\alpha_t,z_{t+1}}}\\
&=\frac{1}{\rho}\PP{A^\alpha_t=a^\alpha_t,Z_{t+1}=z_{t+1}\mid M=m_k,H_t=h_t}\\
&\hspace{3cm}\cdot\PP{M=m_k\mid H_t=h_t}\\
&=\frac{1}{\rho}\PP{A^\alpha_t=a^\alpha_t,Z_{t+1}=z_{t+1}\mid M=m_k,H_t=h_t}p_t(m_k),
\end{align*}
where $\rho$ is some normalization constant. Moreover,
\begin{align*}
\lefteqn{\PP{A^\alpha_t=a^\alpha_t,Z_{t+1}=z_{t+1}\mid M=m_k,H_t=h_t}}\\
&=\PP{Z_{t+1}=z_{t+1}\mid A^\alpha_t=a^\alpha_t,H_t=h_t,M=m_k}\\
&\hspace{3cm}\cdot\PP{A^\alpha_t=a^\alpha_t\mid H_t=h_t,M=m_k}\\
&=\sum_{y\in\X_k}\O_k(z_{t+1}\mid y,a^\alpha_t)\\
&\quad\cdot\PP{X_{t+1}=y\mid A^\alpha_t=a^\alpha_t,H_t=h_t,M=m_k}\pi_t(a^\alpha\mid h_t),
\end{align*}
where the last equality follows from the fact that the agent's action selection given the history does not depend on the task $M$. Therefore,
\begin{align*}
\lefteqn{\PP{A^\alpha_t=a^\alpha_t,Z_{t+1}=z_{t+1}\mid M=m_k,H_t=h_t}}\\
&=\sum_{x,y\in\X_k}\O_k(z_{t+1}\mid y,a^\alpha_t)
                    \P_k(y\mid x,a^\alpha_t)
                    b_{k,t}(x)\pi_t(a^\alpha\mid h_t).
\end{align*}
Putting everything together, we get
\begin{multline}\label{Eq:Task-update}
p_{t+1}(m_k)
=\frac{1}{\rho}\sum_{x,y\in\X_k}\O_k(z_{t+1}\mid y,a^\alpha_t)\\
  \cdot \P_k(y\mid x,a^\alpha_t) b_{k,t}(x)\pi_t(a^\alpha\mid h_t)p_t(m_k),
\end{multline}
with
\begin{multline*}
\rho=\sum_{k=1}^K\sum_{x,y\in\X_k}\O_k(z_{t+1}\mid y,a^\alpha_t)\\
  \cdot\P_k(y\mid x,a^\alpha_t)b_{k,t}(x)\pi_t(a^\alpha\mid h_t)p_t(m_k).
\end{multline*}

Note that the update \eqref{Eq:Task-update} requires the ad hoc agent to keep track of the beliefs $b_{k,t}$ for the different POMDPs $\hat{m}_k$. In other words, at each time step $t$, upon executing its individual action $a^\alpha_t$ and observing $z_{t+1}$, the agent updates each belief $b_{k,t}$ using \eqref{Eq:Belief-update}, yielding
\begin{equation}\label{Eq:Belief-update-2}
b_{k,t+1}(y)=\frac{1}{\rho}\sum_{x\in\X_k}b_{t,k}(x)\P_k(y\mid x,a^\alpha_t)\O_k(z_{t+1}\mid y,a^\alpha_t),
\end{equation}
where $\rho$ is the corresponding normalization constant. Since some of the computations in the update \eqref{Eq:Task-update} are common to the update \eqref{Eq:Belief-update-2}, some computational savings can be achieved by caching the intermediate values.

Finally, at each time step $t$, we define the {\em loss} for selecting an action $a^\alpha\in\A^\alpha$ when the target task is $m_k$ to be
\begin{equation}
\ell_t(a^\alpha\mid m_k)=v^{\hat{\pi}_k}(b_{k,t})-q^{\hat{\pi}_k}(b_{k,t},a^\alpha),
\end{equation}
where $\hat{\pi}_k$ is the solution to the POMDP $\hat{m}_k$. 

It is important to note that both $v^{\hat{\pi}_k}(b_{k,t})$ and $q^{\hat{\pi}_k}(b_{k,t},a^\alpha)$ can be computed from $\hat{v}^{\hat{\pi}_k}$ at runtime, while the latter can be computed offline when solving the POMDP $\hat{m}_k$. Note also that $\ell_t(a^\alpha\mid m_k)\geq 0$ for all $a^\alpha$, and $\ell_t(a^\alpha\mid m_k)=0$ only if $\hat{\pi}_k(a^\alpha\mid b_{k,t})>0$. 


\subsection{Loss bounds for ATPO}

We conclude this section by providing a bound on the loss incurred by our approach. The bound can be derived from a standard compression lemma from \citet{banerjee06icml}, by comparing the performance of an agent using an arbitrary constant distribution over tasks. Let $p$ denote a distribution over $\M$, and define
\begin{align*}
L_t(p)
  &=\sum_{k=1}^Kp(m_k)\ell_t(\hat{\pi}_k\mid m^*),
\end{align*}
where $m^*$ is the (unknown) target task, and
\begin{equation*}
\ell_t(\hat{\pi}_k\mid m^*)=\sum_{a^\alpha\in\A^\alpha}\hat{\pi}_k(a^\alpha\mid b_{k,t})\ell_t(a^\alpha\mid m^*).
\end{equation*}
We have the following result.

\begin{proposition}\label{Prop:Bound}
Let $q$ denote an arbitrary stationary distribution over the tasks in $\M$, and $\set{p_t}$ the sequence of beliefs over tasks generated by ATPO. Then,
\begin{equation}\label{Eq:Bound}
\sum_{t=0}^{T-1}L_t(p_t)\leq \sum_{t=0}^{T-1}L_t(q)+\sqrt{\frac{2}{T}}\sum_{t=0}^{T-1}\KL(q\parallel p_t)+\sqrt{\frac{T}{2}}\cdot \frac{R_{\max}^2}{(1-\gamma)^2}.
\end{equation}
\end{proposition}
Due to space limitations, we include the proof in Appendix~\ref{sec:proof}. Unsurprisingly---and aside from the term $\sqrt{\frac{T}{2}}\cdot \frac{R_{\max}^2}{(1-\gamma)^2}$, which is independent of $q$ and grows sublinearly with $T$---the bound in \eqref{Eq:Bound} states that the difference between the performance of ATPO and that obtained using a constant distribution (for example, the distribution concentrated on $m^*$) is similar to those reported by \citet{banerjee06icml} for Bayesian online prediction with bounded loss, noting that 
\begin{equation}
\sum_{t=0}^{T-1}\KL(q\parallel p_t)=\KL(\vec{q}\parallel\vec{p}_{0:T-1}),
\end{equation}
where we write $\vec{q}$ and $\vec{p}_{0:T-1}$ refer to distributions over sequences in $\M^T$.


\begin{figure*}[!tb]
    \begin{subfigure}[t]{0.3\textwidth}
    \centering
    \includegraphics[width=.7\textwidth]{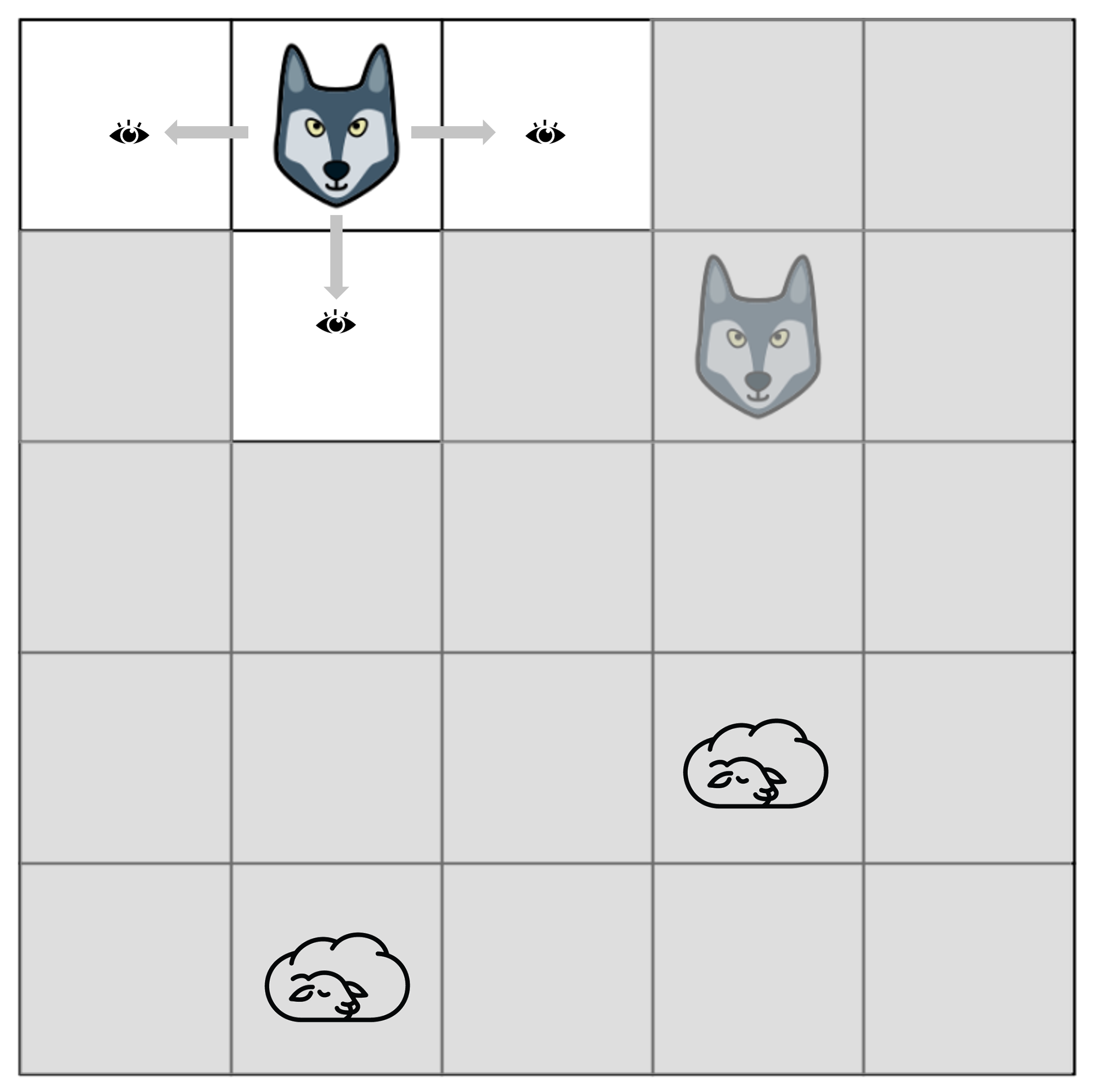}
    \caption{The Night-time Pursuit domain. Two predators must simultaneously capture two sleeping preys.}
    \label{fig:pursuit}
    \end{subfigure}\hfill
    \begin{subfigure}[t]{0.3\textwidth}
    \centering
    \includegraphics[width=.7\textwidth]{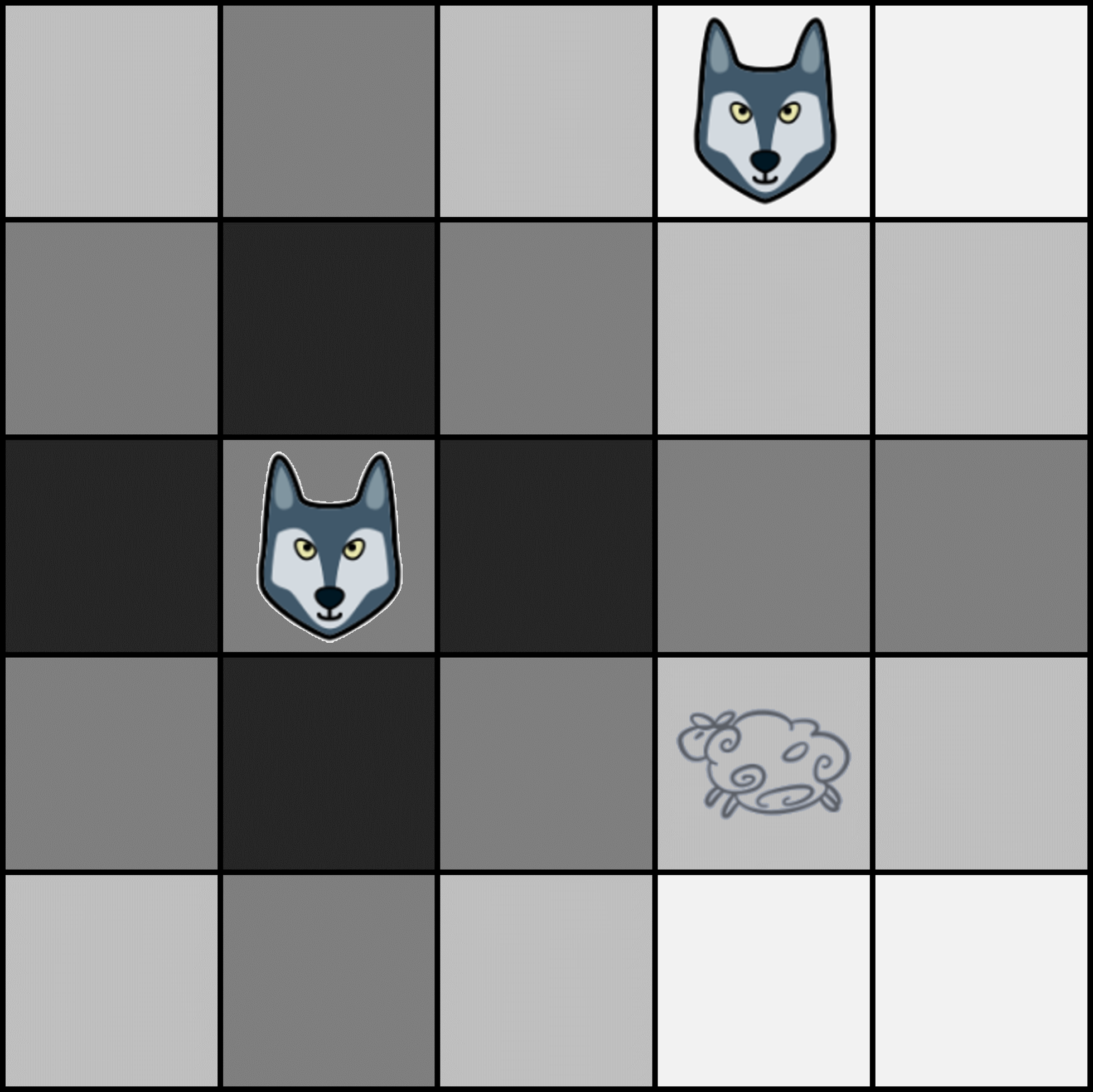}
    \caption{Pursuit under partial observability. Two predators must surround a moving prey. One of the predators has limited perception.}
    \label{fig:pursuit-partial}
    \end{subfigure}\hfill
    \begin{subfigure}[t]{0.3\textwidth}
    \centering
    \includegraphics[width=.7\textwidth]{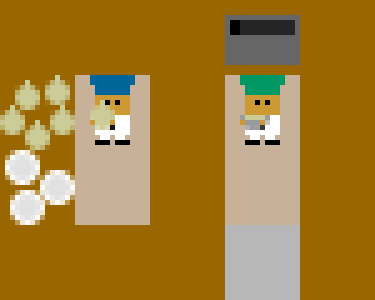}
    \caption{The Overcooked domain \citep{carroll19nips}.}
    \label{fig:overcooked}
    \end{subfigure}
    \caption{The three domains used in the experiments.}
\end{figure*}

\section{Evaluation}%
\label{Sec:Evaluation}

In order to evaluate ATPO, we ran several experiments to respond to the following questions: 
\begin{enumerate}
\item How does ATPO compare with the optimal teammate in terms of task performance? 
\item How does ATPO scale with: (i) the size of the underlying problem? (ii) the number of tasks in $\M$? 
\item How well is ATPO able to deal with the uncertainty in both transitions and observations?
\end{enumerate}
To answer these questions, we assess the performance of ATPO in three different environments, namely the Night-time Pursuit, the Pursuit under partial observability, and the Overcooked domain. Given the easy parameterization of the Night-time Pursuit, we used it to answer all the questions above. Furthermore, we used the other environments to complement the Night-time Pursuit in addressing Question~(1).

\subsection{Test environments}

In our experiments, we consider three distinct environments. 
\begin{itemize}
\item {\bf The Night-time Pursuit domain}. In this environment, two predators must capture two sleeping preys (see Fig.~\ref{fig:pursuit}). The preys do not change their position, and the predators must coordinate regarding which predator captures each prey. The ad hoc agent can only observe the neighboring cells. The position of the preys is not encoded in the state. Rather, the different tasks in $\M$ correspond to different configurations of the sleeping preys. Given its easy parameterization, we use this domain to evaluate the dependence of the performance of the different methods with different environment configurations.
\item {\bf Pursuit under partial observability}. This environment is a modification of the standard Pursuit domains to include only two predators and one moving prey (see Fig.~\ref{fig:pursuit-partial}. The perception of the ad hoc agent deteriorates with the distance to its position: cells further away are observed with higher uncertainty than the cells closer to the agent. The different tasks in $\M$ correspond to different capturing behaviors of the teammate.
\item {\bf The Overcooked domain}. This domain is depicted in Fig.~\ref{fig:overcooked}. In this environment, two cooks must coordinate to fetch the necessary ingredients to produce a soup. The scenario was proposed by \citep{carroll19nips} as a benchmark to evaluate cooperation. However, it does not feature partial observability. The different tasks correspond to different roles that the teammate can assume.
\end{itemize}
We provide detailed descriptions of the different scenarios in Appendix~\ref{Sec:Envs}.


\subsection{Baselines}

We compare ATPO with different baseline agents in each domain, deploying each baseline agent to act as the ad hoc agent. We consider the following baselines:
\begin{itemize}

    \item \textbf{Value Iteration}: The Value Iteration agent knows the target task and can perfectly observe the current state of the environment. It follows the optimal policy for the associated MMDP, computed using value iteration. The performance of this agent can be considered as an upper bound to ATPO.
    \item \textbf{Perseus}: The Perseus agent knows the target task, but suffers from partial observability. It follows the optimal policy for the target POMDP computed using Perseus \citep{spaan2005perseus}. The performance of this agent can also be considered as an upper bound to ATPO.
    \item \textbf{Bayes Online Prediction for Ad hoc teamwork (BOPA)}: The BOPA agent corresponds to the ad hoc algorithm of \citet{ribeiro2021helping}. To handle partial observability, we adopt the most-likely state heuristic, where the state with the highest belief in each task is considered the current state. This baseline cannot be used in scenarios where different tasks have different state spaces.
    \item \textbf{Assistant}: The Assistant agent corresponds to the algorithm of \citet{fern2007decision} for assistive tasks. It can be seen as an ad hoc algorithm where the agent can observe the actions of the teammate.
    \item \textbf{Random Policy}: This agent selects actions randomly. The performance of this agent can be considered as a lower bound to ATPO and the other agents.
\end{itemize}

\subsection{Metrics}

To answer the questions outlined at the beginning of this section, we consider four different metrics. First, {\em Effectiveness}, measuring whether ATPO can identify the correct task. Second, {\em Efficiency}, measuring whether an agent can solve the task in near-optimal time. Third, {\em Robustness}, assessing the performance of the agent as a function of the noise in the environment, $\eps$.\footnote{A higher noise increases the uncertainty in both state transitions and observations.} Finally, {\em scalability}, measuring the dependence of the agent's performance on the problem size and the number of tasks in $\M$.

\begin{figure}[!tb]
    \begin{subfigure}[t]{0.42\textwidth}
    \centering
    \includegraphics[width=\textwidth]{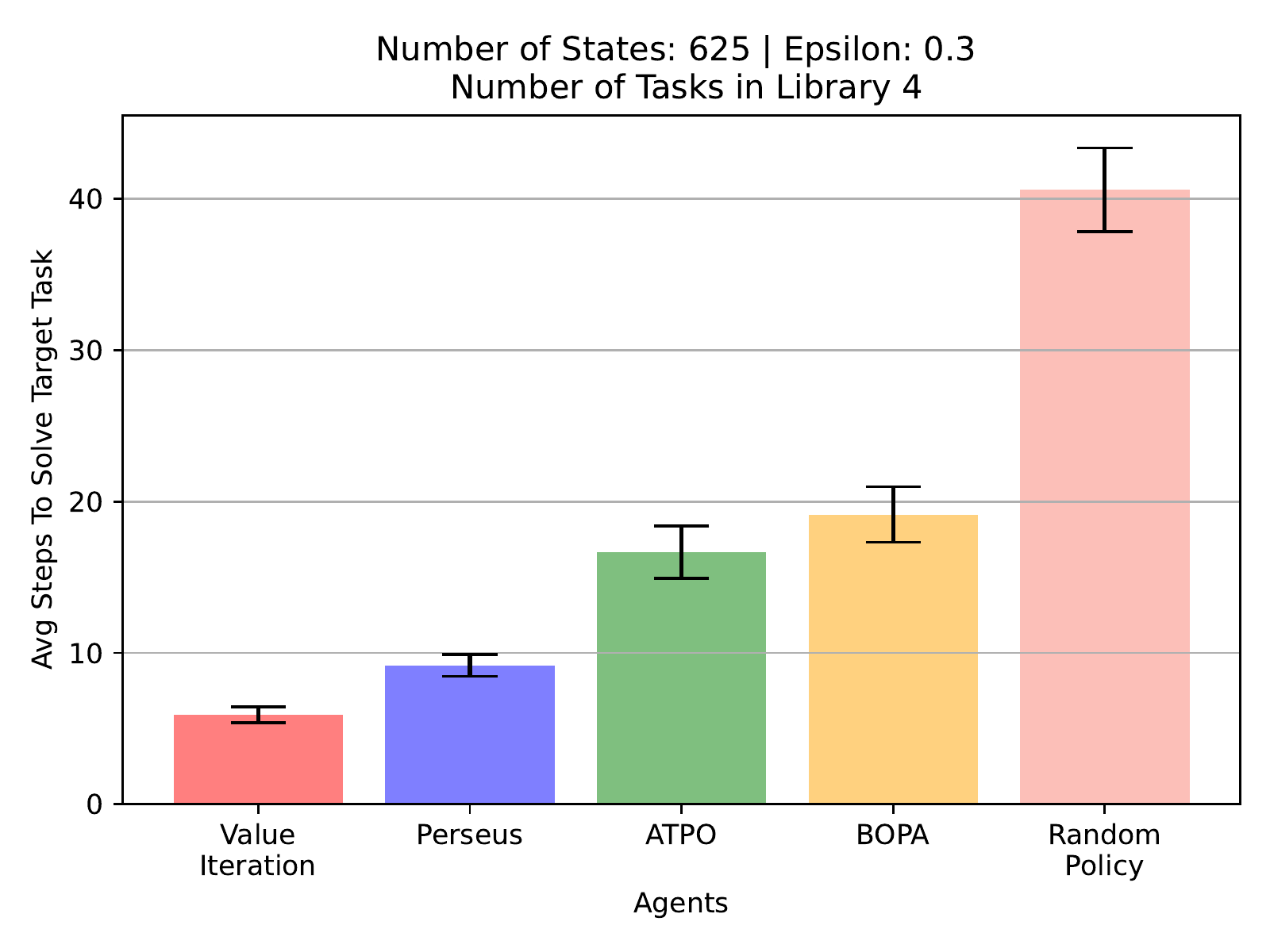}
    \caption{Average steps to solve a task for each agent. }
    \label{fig:pursuit:baseline-results}
    \end{subfigure}\hfill
    \begin{subfigure}[t]{0.42\textwidth}
    \centering
    \includegraphics[width=\textwidth]{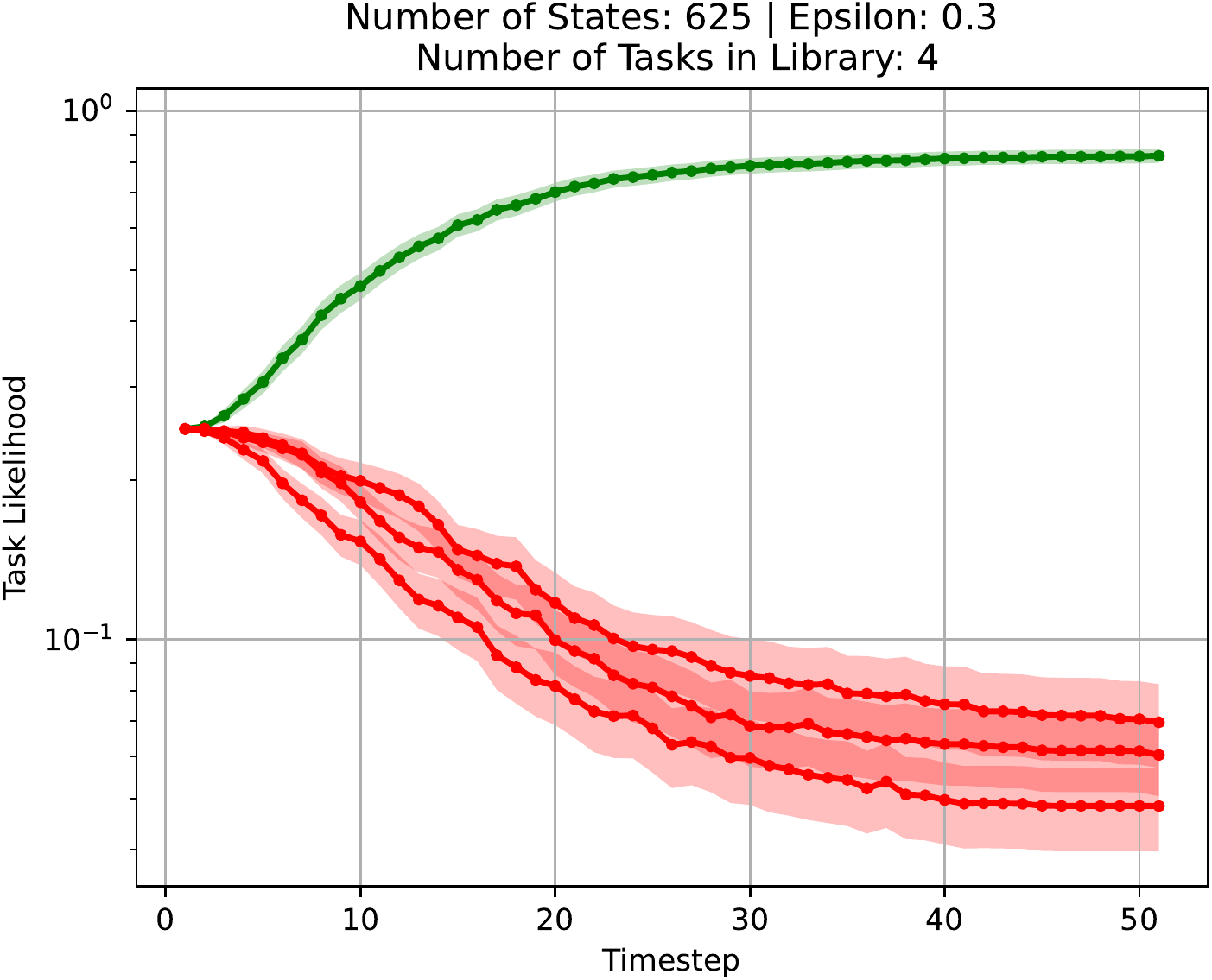}
    \caption{Task likelihood averaged over all trials. The likelihood of the correct task is marked in green, while those of other tasks are marked in red.}
    \label{fig:pursuit:baseline-entropy}
    \end{subfigure}\hfill
    \caption{Performance in the baseline configuration for the Night-time Pursuit domain.}
    \label{Fig:Results-pursuit}
\end{figure}

\begin{figure*}[!tb]
    \begin{subfigure}[t]{0.33\textwidth}
    \centering
    \includegraphics[width=\textwidth]{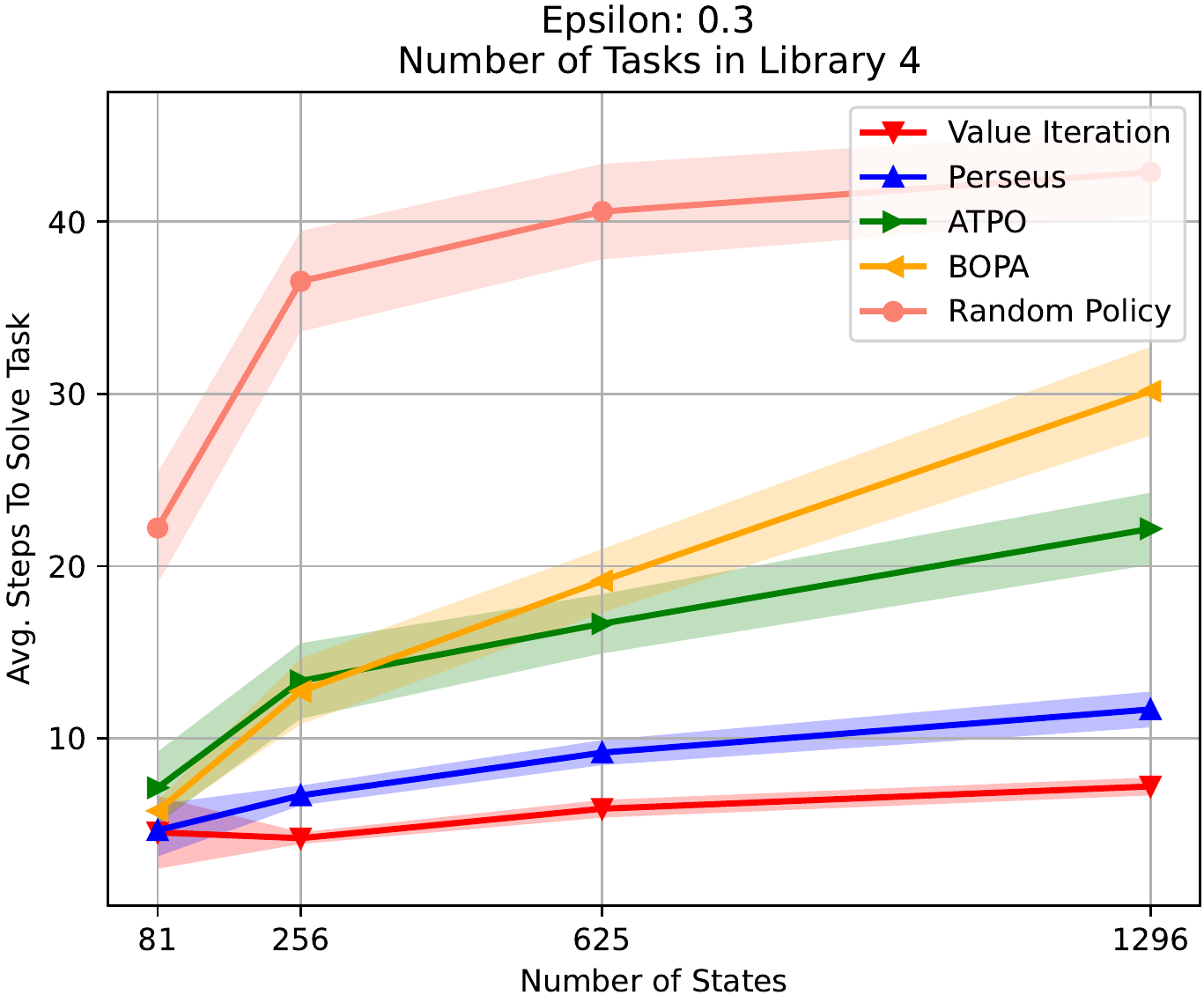}
    \caption{Task performance as a function of $\abs{\X}$.}
    \label{fig:pursuit:scalability-states}
    \end{subfigure}\hfill
    \begin{subfigure}[t]{0.33\textwidth}
    \centering
    \includegraphics[width=\textwidth]{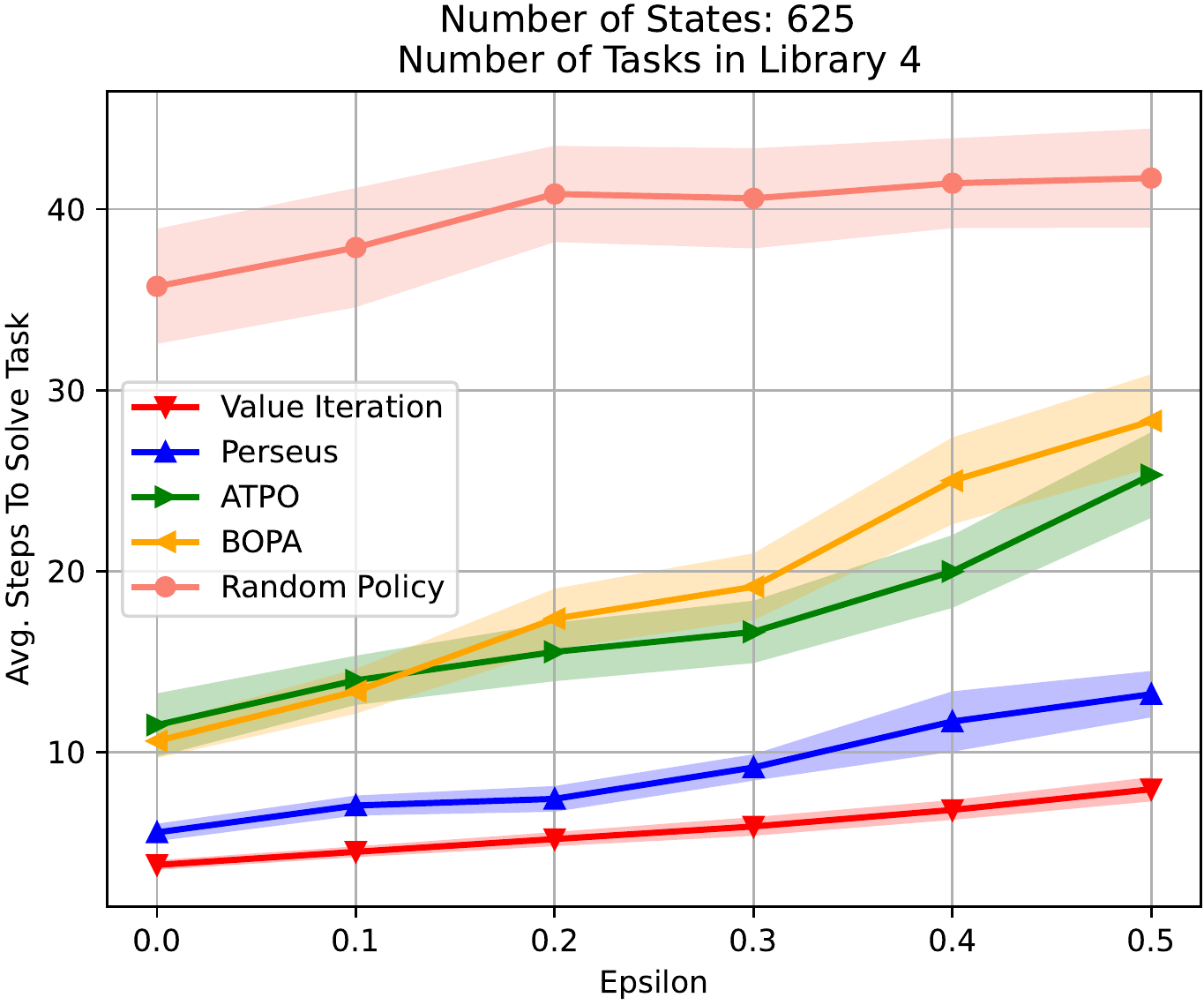}
    \caption{Task performance as a function of $\eps$.}
    \label{fig:pursuit:scalability-epsilon}
    \end{subfigure}\hfill
    \begin{subfigure}[t]{0.33\textwidth}
    \centering
    \includegraphics[width=\textwidth]{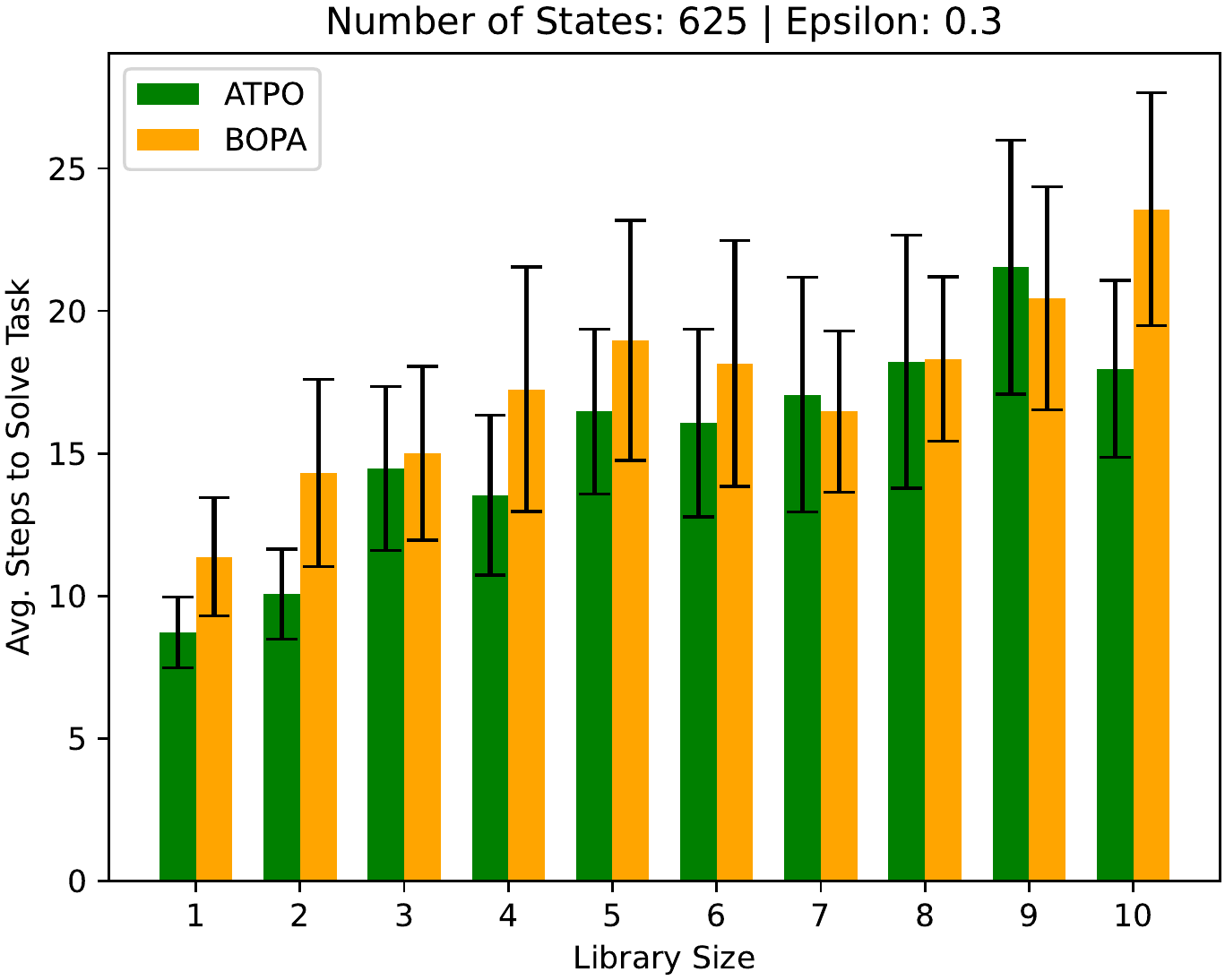}
    \caption{Task performance as a function of $\abs{\M}$.}
    \label{fig:pursuit:robustness}
    \end{subfigure}
    \caption{Results on the Night-time Pursuit domain, as we vary $|\X|$, $\eps$ and $|\M|$ around their baseline values.}
    \label{Fig:Results-pursuit}
\end{figure*}

In all experiments, the reported values consist of averages and $95\%$ confidence intervals over $32$ independent trials, where a single trial consists of running on a given task over a finite horizon. We consider a task from the Pursuit domains complete whenever the interaction reaches a horizon of 50 timesteps and a task from the Overcooked domain complete whenever the interaction reaches a horizon of 75 timesteps.


\section{Results}

We now discuss the results of our experiments. Due to its simplicity and ease of interpretation, we use the Night-time pursuit domain for a more extensive evaluation of the different approaches. We then complete our evaluation by reporting the comparative performance of ATPO in the other two domains.

\subsection{Night-time Pursuit Domain}

In this scenario, our baseline environment configuration is such that $|\X|=625$, $\eps=0.30$ and $|\M|=4$. For each of the possible tasks in $\M$, we run 32 independent trials. Figure \ref{fig:pursuit:baseline-results} plots, by agent, the number of steps it takes each agent to solve the target task and ATPO's belief over tasks, averaged over the 128 trials and with a $95\%$ confidence intervals.

As expected, Value Iteration achieved the best results. Conversely, the random policy achieved the worst results. Our approach, ATPO, stands between Perseus and BOPA agents because Perseus has more information (i.e., it knows the target task) and BOPA discards important history information by considering only the most likely state. We also observe that ATPO can pinpoint the target task quite efficiently.

We then compare the performance of the different ad hoc agents as we vary each of the three environment parameters around the baseline configuration. We vary $\abs{\X}$ between $81$ states (in a $3\times 3$ world) and $1,296$ states (in a $6\times 6$ world), $\eps$ between $0.0$ and $0.5$, and $\abs{\M}$ between 1 and 10. We run 32 independent trials for each configuration, where the task library and target task are randomly selected from a pool of 32 possible tasks. The results are depicted in Fig.~\ref{Fig:Results-pursuit}. 

The results also correspond to what we expected. In terms of task performance (Fig.~\ref{fig:pursuit:scalability-states}), all methods scale gracefully as the problem size increases. In particular, the relative performance of all methods remains mostly the same, with some methods (such as BOPA) showing larger degradation than the other methods due to the higher state uncertainty expected as the state space increases. Similar observations can be made regarding the dependence on $\eps$ (Fig.~\ref{fig:pursuit:scalability-epsilon}). Once again, BOPA presents larger degradation because---as the noise increases---it is less capable of tracking the underlying state. In both cases, the performance of ATPO is slightly worse than the two more informed agents (Perseus and Value Iteration) and better than the other baselines (BOPA and random agents).

Finally, when analyzing the dependence on $\abs{\M}$, we compare only BOPA and ATPO, since these are the only two algorithms that consider the existence of multiple tasks. As observed in Fig.~\ref{fig:pursuit:robustness}, the performance of both algorithms shows some slight decay as $\abs{\M}$ increases, since it will take more observations for the ad hoc agent to identify the target task.

\begin{figure}[!tb]
    \begin{subfigure}[t]{0.33\textwidth}
    \centering
    \includegraphics[width=\textwidth]{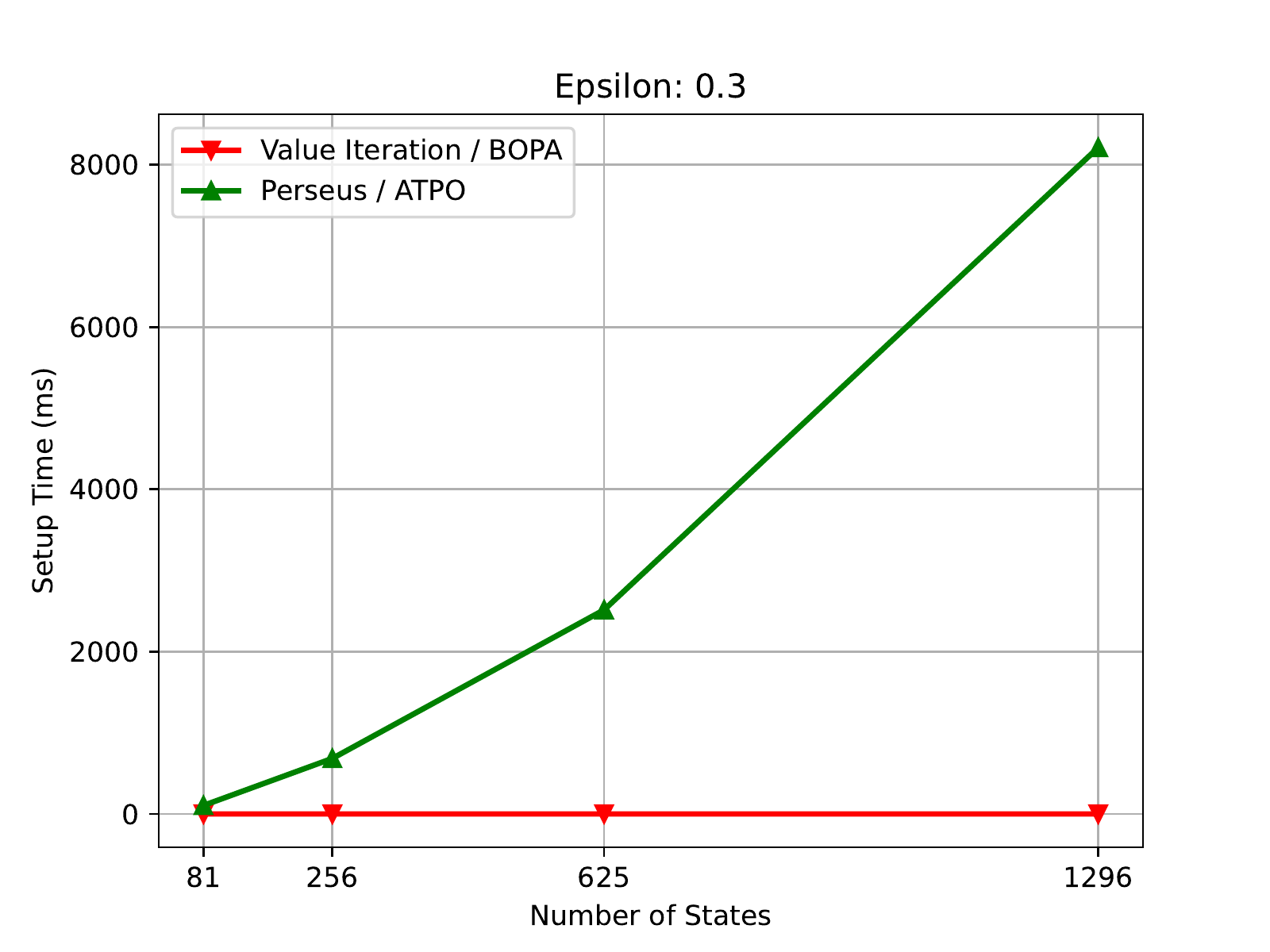}
    \caption{Total setup time for the different agents for a single task (offline). Results correspond to a single instantiation of each agent.}
    \label{fig:pursuit:setup}
    \end{subfigure}\hfill
    \begin{subfigure}[t]{0.33\textwidth}
    \centering
    \includegraphics[width=\textwidth]{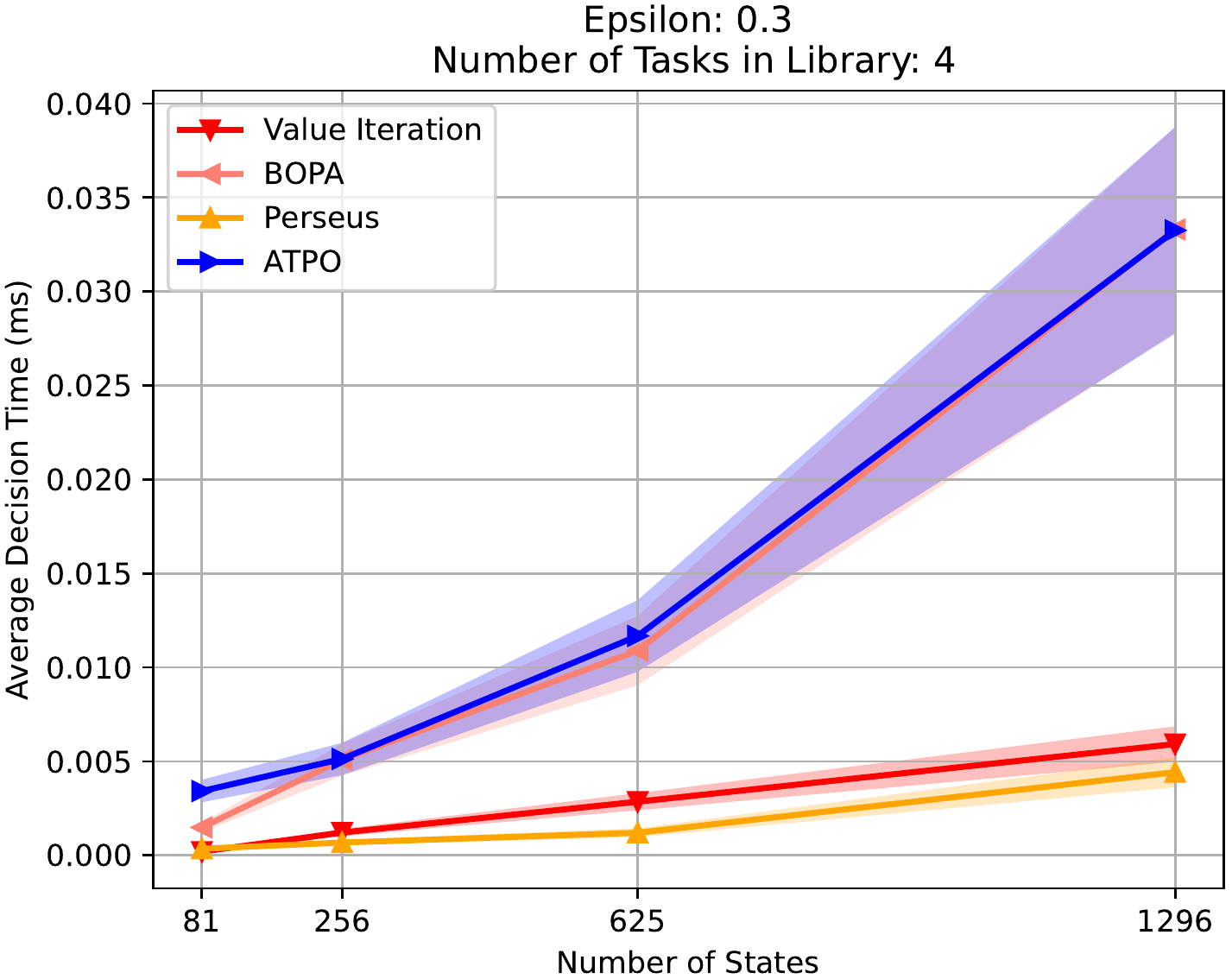}
    \caption{Average decision time per timestep for each agent. Results correspond to 150 samples and are reported with a confidence of $95\%$.}
    \label{fig:pursuit:decision}
    \end{subfigure}\hfill
    \caption{Computation time as a function of the problem size for the Night-time Pursuit domain.}
    \label{Fig:Computation-pursuit}
\end{figure}

To conclude our analysis of the Night-time Pursuit domain, we measured the online and offline computation effort required by the different methods. The results are depicted in Fig.~\ref{Fig:Computation-pursuit}.

As seen in Fig.~\ref{fig:pursuit:setup}, the setup times (corresponding to offline computation) are significantly different between the approaches relying on MDP formulations (value iteration and BOPA) and those relying on POMDP formulations (ATPO and Perseus). Value iteration and BOPA show a linear increase in computation time as the size of the task grows. ATPO and Perseus, on the other hand, show a super-linear growth as a function of $\abs{\X}$. Consequently, for the larger domains, the computation time of the POMDP-based methods is several orders of magnitude larger than that of MDP-based methods. Note also that the computation times reported are for a single task, meaning that the actual computation times for BOPA and ATPO should be multiplied by $\abs{\M}$. These results are unsurprising, as POMDPs are known to be significantly harder to solve than their fully observable counterparts \citep{spaan05jair}.

Conversely, Fig.~\ref{fig:pursuit:decision} depicts the decision time (corresponding to online computation). We can now observe that the increase in computation time grows much more gracefully. As expected, the two methods that identify the target task (BOPA and ATPO) exhibit larger decision times than those that know the target task (Perseus and value iteration).

\subsection{Pursuit Domain under Partial Observability}

\begin{figure}[!tb]
    \begin{subfigure}[t]{0.33\textwidth}
    \centering
    \includegraphics[width=\textwidth]{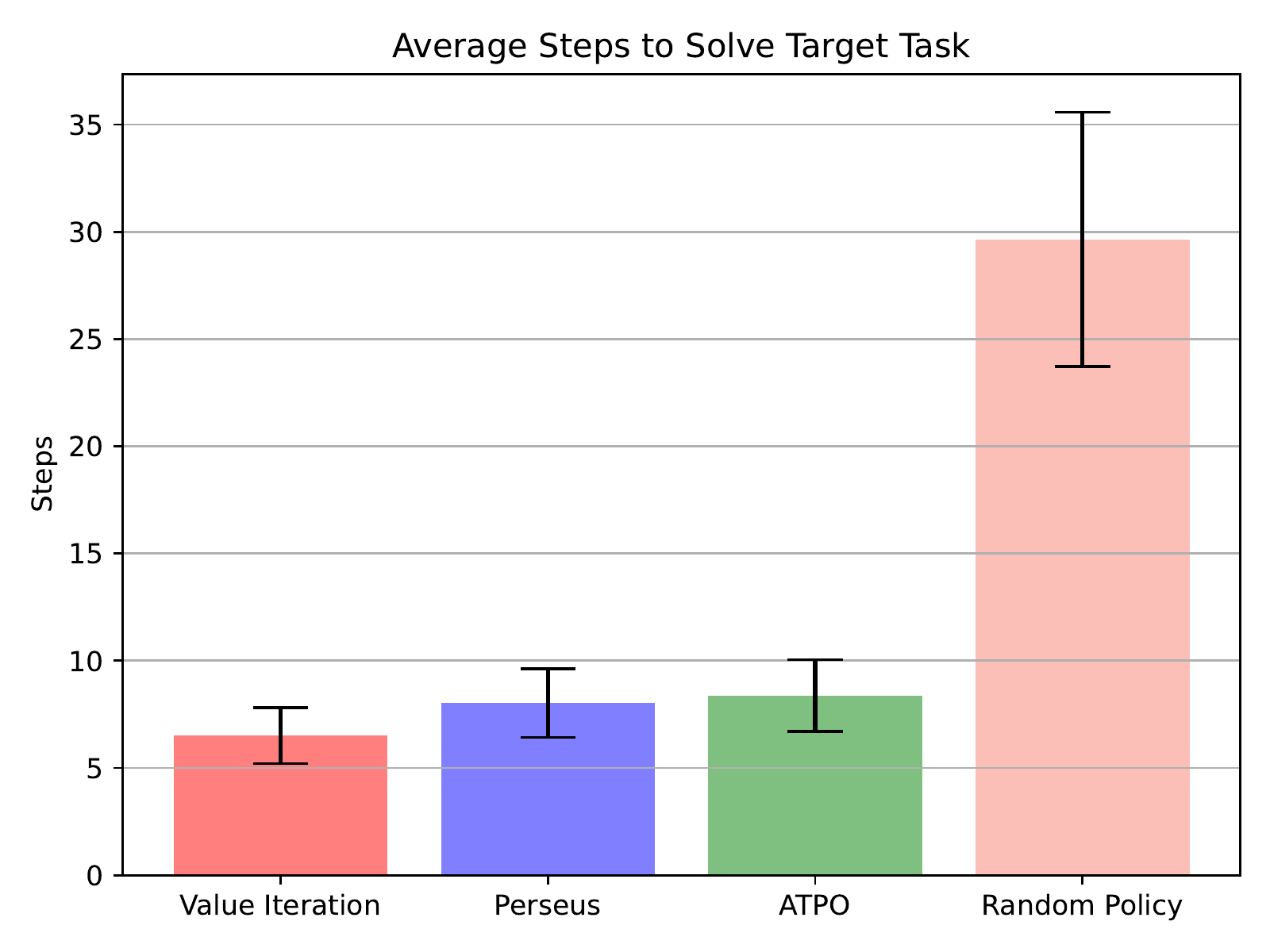}
    \caption{Average steps to solve a task for each agent on the Pursuit domain under partial observability. Results correspond to 96 total trials and are reported with a confidence of $95\%$.}
    \label{fig:pursuit-partial:baseline-results}
    \end{subfigure}\hfill
    \begin{subfigure}[t]{0.33\textwidth}
    \centering
    \includegraphics[width=\textwidth]{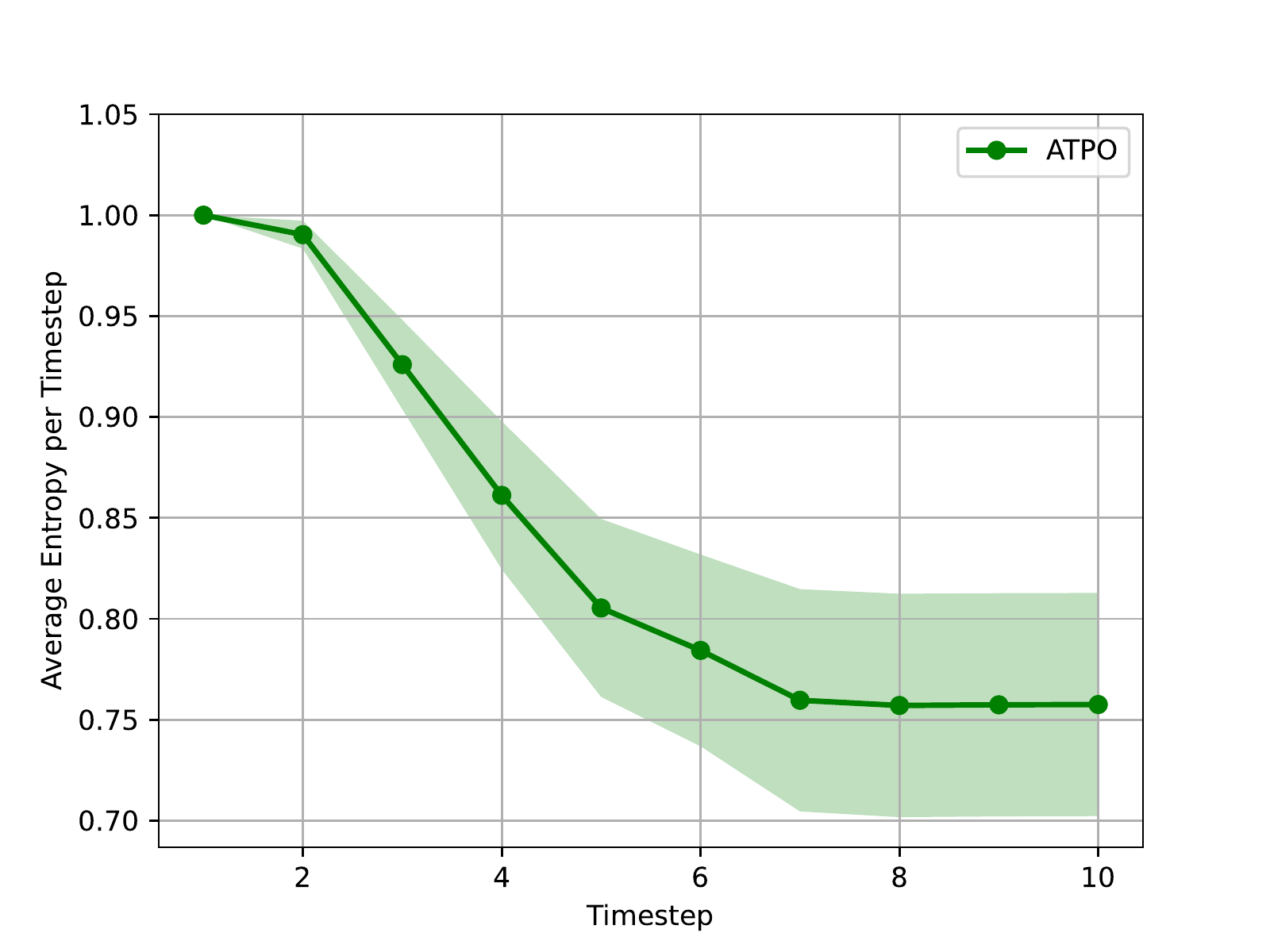}
    \caption{Average entropy per timestep from ATPO model priors $p(m_k)$ (lower scale). Results correspond to 96 total trials and are reported with a confidence of $95\%$.}
    \label{fig:pursuit-partial:baseline-entropy}
    \end{subfigure}\hfill
    \caption{Performance in Pursuit under partial observability.}
    \label{Fig:Results-pursuit-partial}
\end{figure}

We now use the Pursuit domain under partial observability to complement the Night-time Pursuit in addressing Question~(1) in Section~\ref{Sec:Evaluation}. We consider a fixed configuration, where $|\X|=552$, corresponding to a $5\times5$ world, $\epsilon(d)=1-0.15d$, and $|\T|=3$. The different tasks correspond to different teammate behaviors, for which we consider the following teammates:
\begin{itemize}
    \item \textbf{Greedy teammate}: Always moves in the direction of the prey, regardless of any obstacles along the way.
    \item \textbf{Teammate-aware teammate}: Computes the shortest path to the prey using A* search, taking into account the position of the ad hoc agent.
    \item \textbf{Probabilistic-destinations teammate}: Tries to surround the prey, taking into account the position of the ad hoc agent.
\end{itemize}
For each of the three possible teammate types, we ran 32 independent trials (adding to a total of 96 trials). The results are depicted in Fig.~\ref{Fig:Results-pursuit-partial}.

The results are similar to those observed in Night-time Pursuit. Value Iteration is again the best performing agent, and the random policy the worst performing one. Another noteworthy aspect is that there is almost no difference between Perseus and ATPO, indicating the near-optimal performance of the latter. We can also observe, in Fig.~\ref{fig:pursuit-partial:baseline-entropy}, that the uncertainty regarding the target task slowly decreases, again showing that ATPO can effectively identify the target task (in this case, the type of teammate).

\subsection{Overcooked Domain}

\begin{figure}[!tb]
    \begin{subfigure}[t]{0.42\textwidth}
    \centering
    \includegraphics[width=\textwidth]{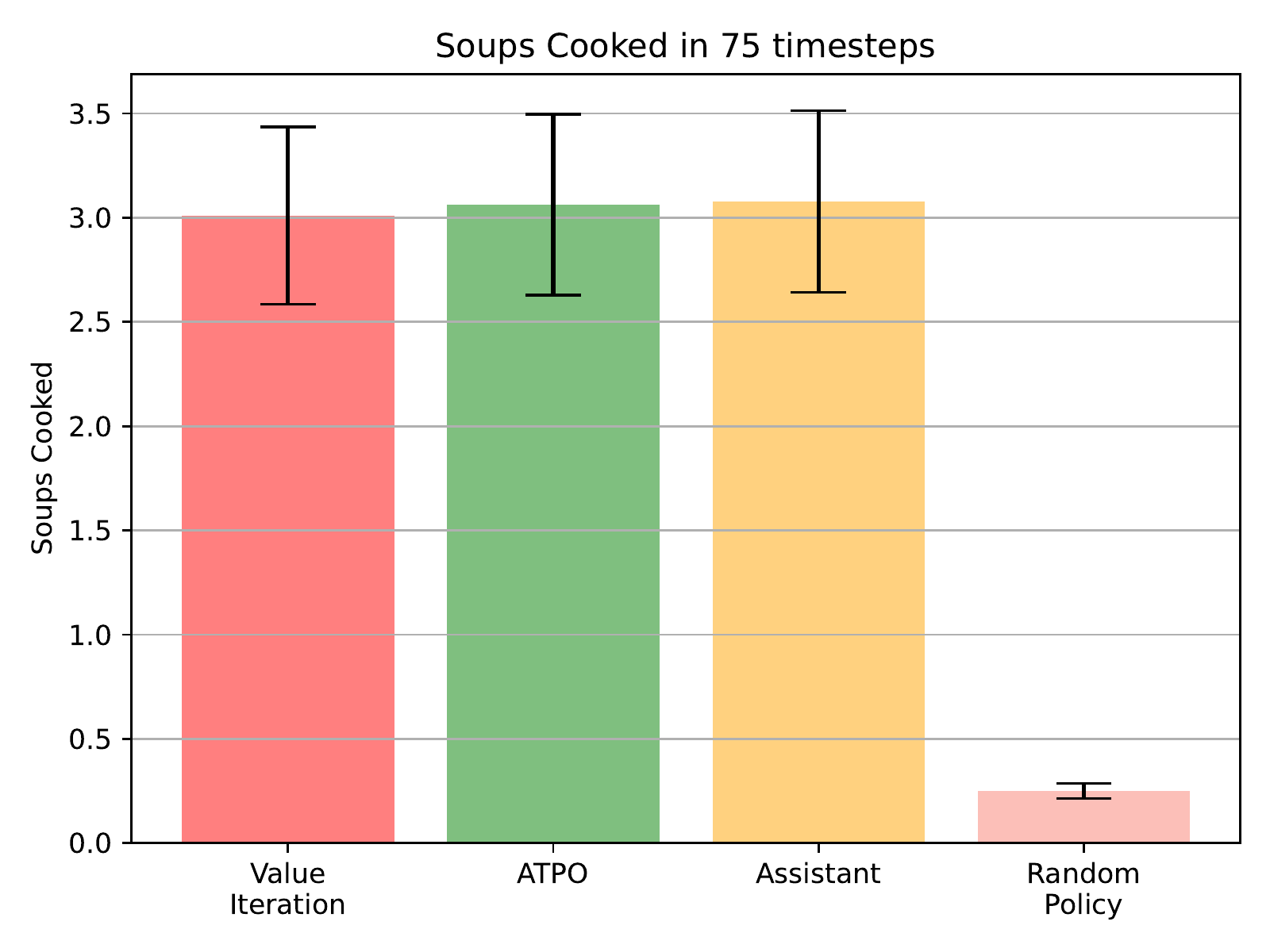}
    \caption{Average number of soups cooked in a 75 timestep horizon. Results correspond to 192 total trials and are reported with a confidence of $95\%$.}
    \label{fig:overcooked:baseline}
    \end{subfigure}\hfill
    \begin{subfigure}[t]{0.42\textwidth}
    \centering
    \includegraphics[width=\textwidth]{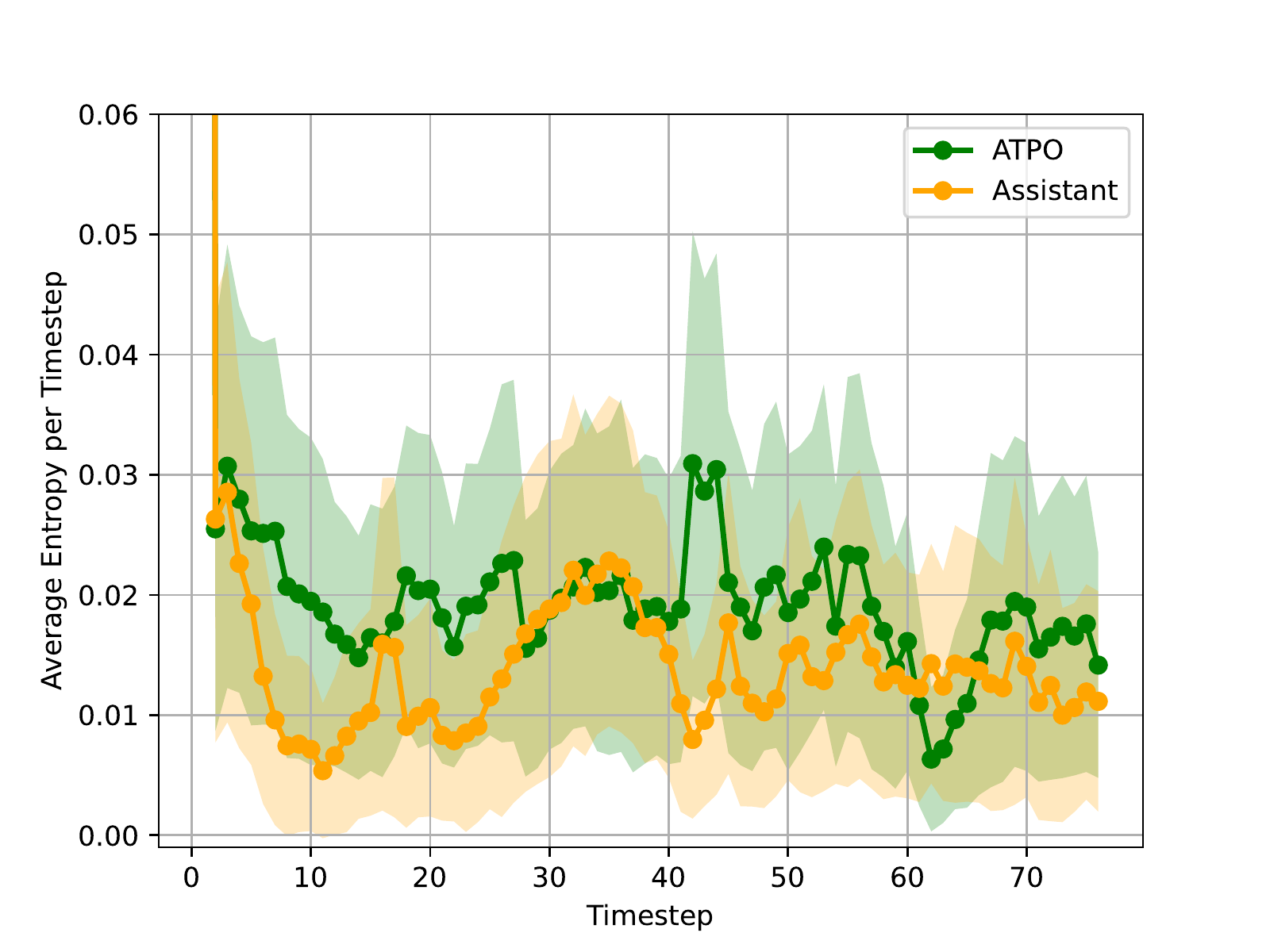}
    \caption{Average entropy per timestep from ATPO's and The Assistant's model priors $p(m_k)$ (lower scale). Results correspond to 192 total trials and are reported with a confidence of $95\%$.}
    \label{fig:overcooked:baseline-entropy-closeup}
    \end{subfigure}\hfill
    \caption{Performance in Overcooked.}
    \label{Fig:Results-overcooked}
\end{figure}

Finally, we consider the third test scenario---the Overcooked domain. This scenario has been proposed as a benchmark to evaluate cooperation. Therefore, it is an interesting domain in the context of ad hoc teamwork. Unlike the previous scenarios, however, Overcooked does not have partial observability. In our experiments, we evaluate the performance of ATPO when the ad hoc agent can be either a helper (against four possible types of cook teammate) or a cook (against two possible types of helper teammate). The possible teammates with a cook's role are Greedy, Dummy, and Upper and Downer. Furthermore, teammates with a helper role are Greedy and Dummy. Greedy agents try to complete the task as quickly as possible, not waiting for the ad hoc agent. Dummy agents only act as a response to the actions of the ad hoc agent. Upper and Downer agents perform the cook role but without moving---remaining only in the upper or lower cell, respectively. For each of the 6 possible tasks, we ran 32 independent trials, adding up to 192 total independent trials. Results are shown in Fig.~\ref{Fig:Results-overcooked}.

By looking at the number of soups cooked (Fig.~\ref{fig:overcooked:baseline}), we can observe that the random policy is only able to finish a small number of soups---mostly by chance. The other three methods perform similarly, as there is no statistical significance between the observed differences. ATPO thus performs optimally---a performance that is indistinguishable from that of value iteration. It is also similar to that of the Assistant algorithm. 

We can also observe in Fig.~\ref{fig:overcooked:baseline-entropy-closeup} the entropy of the belief $p_t$ at each time step $t$, both for ATPO and the Assistant. Both agents can identify the target task, on average, in less than 10 timesteps. We can also see no statistically significant difference between both agents, even though the Assistant was able to achieve a slightly lower entropy overall when compared to ATPO. This result is not unexpected, given that the Assistant can fully observe the teammate's actions. Nevertheless, our results show that both agents are effective in identifying the target task with great certainty, but also that our approach, ATPO, can do so without being able to observe the teammate's actions.


\section{Conclusion}

This paper presents and evaluates the Ad Hoc Teamwork under Partial Observability (ATPO) algorithm, a novel approach for the ad hoc teamwork problem. In this setting, an agent has to learn to cooperate with teammates in solving an unknown task, without observing the current state of the environment, the teammates' actions, and the environment's reward signals.

Unlike previous approaches that assume relevant information is always readily accessible to the agent (e.g., environment's reward signals, teammate's actions, and state observations), ATPO does not rely on the assumption that the multi-agent system is fully observable, relying only on partial observations. Our results in three domains --- the Night Time Pursuit domain, the Pursuit domain under partial observability, and the Overcooked domain --- show that our approach effectively identifies which task is being performed from a pool of possible tasks. The results also show that it is: (i) efficient at solving the correct task in near-optimal times, as if it knew the task beforehand; (ii) scalable, by being able to adapt to faulty sensors, actuators, and larger problem sizes; (iii) robust, by being able to adapt to an increasingly larger number of possible tasks.






\bibliographystyle{ACM-Reference-Format} 
\bibliography{sample}


\clearpage

\appendix

\section{Proof of Proposition~\ref{Prop:Bound}}
\label{sec:proof}

We can derive the bound for the loss in \eqref{Eq:Bound} incurred by our agent, when compared against an agent considering a constant distribution $q$ over tasks. We use the following compression lemma from \cite{banerjee06icml}.

\begin{lemma}\label{Lemma:Compression}
Given a set of hypothesis $\H=\set{1,\ldots,H}$, for any measurable function $\phi:\H\to\IR$ and any distributions $p$ and $q$ on $\H$,
\begin{equation}
\EE[h\sim q]{\phi(h)}-\log\EE[h\sim p]{\exp(\phi(h))}\leq\KL(q\parallel p).
\end{equation}
\end{lemma}

We want to bound the loss incurred by our agent after $T$ time steps. Before introducing our result, we require some auxiliary notation. Let $m^*$ denote the (unknown) target task at time step $t$. The expected loss of our agent at time step $t$ is given by
\begin{align*}
L_t(\pi_t)
  &=\EE{\ell_t(A^\alpha\mid m^*)}\\
  &=\sum_{a^\alpha\in\A^\alpha}\pi_t(a^\alpha)\ell_t(a^\alpha\mid m^*)\\
  &=\sum_{k=1}^Kp_t(m_k)\sum_{a^\alpha\in\A^\alpha}\hat{\pi}_k(a^\alpha\mid b_{k,t})\ell_t(a^\alpha\mid m^*)\\
  &=\sum_{k=1}^Kp_t(m_k)\ell_t(\hat{\pi}_k\mid m^*),
\end{align*}
where, for compactness, we wrote
\begin{equation}
\ell_t(\hat{\pi}_k\mid m^*)=\sum_{a^\alpha\in\A^\alpha}\hat{\pi}_k(a^\alpha\mid b_{k,t})\ell_t(a^\alpha\mid m^*).
\end{equation}

Let $q$ denote an arbitrary distribution over $\M$, and define
\begin{equation}
L_t(q)=\sum_{k=1}^Kq(m_k)\ell_t(\hat{\pi}_k\mid m^*).
\end{equation}
Then, setting $\phi(m_k)=-\eta\ell_t(\hat{\pi}_{k_t}\mid m^*)$, for some $\eta>0$, and using Lemma~\ref{Lemma:Compression}, we have that
\begin{equation}
\EE[m\sim q]{\phi(m)}-\log\EE[m\sim p_t]{\exp(\phi(m))}\leq\KL(q\parallel p_t)
\end{equation}
which is equivalent to
\begin{equation}\label{Eq:Ineq1}
-\log\EE[m\sim p_t]{\exp(\phi(m))}\leq \eta L_t(q)+\KL(q\parallel p_t).
\end{equation}
Noting that $-2\eta\frac{R_{\max}}{1-\gamma}\leq\phi(m)\leq 0$ and using Hoeffding's Lemma,%
\footnote{Hoeffding's lemma states that, given a real-valued random variable $X$, where $a\leq X\leq b$ almost surely, 
\begin{equation}
\EE{e^{\lambda X}}\leq\exp\left(\lambda\EE{X}+\frac{\lambda^2(b-a)^2}{8}\right),
\end{equation}
for any $\lambda\in\IR$. 
}
we have that
\begin{equation}\label{Eq:Ineq2}
-\log\EE[m\sim p_t]{\exp(\phi(m))}\geq\eta L_t(p_t)-\frac{\eta^2R_{\max}^2}{2(1-\gamma)^2}.
\end{equation}
Combining \eqref{Eq:Ineq1} and \eqref{Eq:Ineq2}, yields
\begin{equation}
L_t(p_t)\leq L_t(q)+\frac{1}{\eta}\KL(q\parallel p_t)+\frac{\eta R_{\max}^2}{2(1-\gamma)^2}
\end{equation}
which, summing for all $t$, yields
\begin{equation}
\sum_{t=0}^{T-1}L_t(p_t)\leq \sum_{t=0}^{T-1}L_t(q)+\frac{1}{\eta}\sum_{t=0}^{T-1}\KL(q\parallel p_t)+\frac{T\eta R_{\max}^2}{2(1-\gamma)^2}.
\end{equation}
Since $\eta$ can be selected arbitrarily, setting $\eta=\sqrt{\frac{T}{2}}$ we finally get
\begin{equation}
\sum_{t=0}^{T-1}L_t(p_t)\leq \sum_{t=0}^{T-1}L_t(q)+\sqrt{\frac{2}{T}}\sum_{t=0}^{T-1}\KL(q\parallel p_t)+\sqrt{\frac{T}{2}}\cdot \frac{R_{\max}^2}{(1-\gamma)^2} \square
\end{equation}

\section{Environment description}%
\label{Sec:Envs}

We now provide detailed descriptions of our three test scenarios.

\subsection{Night-time pursuit}

Our first evaluation testbed---the Night-time Pursuit domain---re\-pre\-sents a modified version of the Pursuit domain,\footnote{\url{https://www.cs.cmu.edu/afs/cs/usr/pstone/public/papers/97MAS-survey/node8.html}} in which we introduce partial observability. Unlike the traditional Pursuit domain, where $N$ predator agents much capture a moving prey, in the Night-time Pursuit domain, $N$ predator agents must each capture its own sleeping, static prey. In the Night Time Pursuit domain, the preys positions are fixed and not encoded in the state, only in the reward. Different possible configurations for the preys thus correspond to the different tasks in $\M$.

In this scenario, the ad hoc agent can move up, down, left and right, or stay in its current cell. Each moving action succeeds with probability $1-\eps$, for some $\eps\geq 0$, except if there is a wall in the corresponding direction, in which case the position of the agent remains unchanged. When an action fails, the position of the agent remains unchanged.

The ad hoc agent can only observe the neighboring cells. For example, in Fig.~\ref{fig:pursuit}, the ad hoc agent (in the top left corner) can only observe, at each time step, the elements (teammate, preys) in the white cells. {\em It cannot observe its current position}. Observations are also not flawless: whenever an element (teammate, prey) is in a neighboring cell, there is a probability $\eps$ that the agent will fail to observe it.

We model each task $k$ in the Night-time Pursuit domain as a POMDP $(\X,\A^\alpha,\Z,\set{\P_{a^\alpha},a^\alpha\in\A^\alpha},\set{\O_{a^\alpha},a^\alpha\in\A^\alpha},r_k,\gamma)$. A state $x\in\X$ contains information regarding the positions of both agents, and is therefore represented as a tuple $x=(c_1, r_1, c_2, r_2)$, where $(c_n, r_n)$ represents the cell (column and row) where agent $n$ is located. 

Each observation $z\in\Z$ are also represented by a tuple $z=(\hat{u}, \hat{d}, \hat{l}, \hat{r})$, where each entry represents what is observed, respectively, above, below, to the left, and to the right of the agent. For each entry there are three possible values: \textit{Nothing}, \textit{Teammate}, \textit{Prey}. The action space $\A^\alpha$ contains five possible actions, \textit{Up, Down, Left, Right} and \textit{Stay}. The transition probabilities $\set{\P_{a^\alpha},a^\alpha\in\A^\alpha}$ map a state $x$ and action $a^\alpha$ to every possible next state $x'$, taking into account the probability $\epsilon$ of the agent failing to move (considering that the teammate actions always succeed and move the teammate towards its closest prey). Similarly, the observation probabilities $\set{\O_a^\alpha,a^\alpha\in\A^\alpha}$ map a state $x$ and previous action $a^\alpha$ to every an observation $z$, taking into account the probability $\epsilon$ of the agent failing to observe any given element. The reward function $r_k$ assigns the reward of $-1$ for all time steps except those where the preys have been captured, in which case it assigns a reward of $100$ for the one the preys are captured in and $0$ afterwards (in other words, since where dealing with an horizon of 50 steps, we do not reset the environment whenever the preys have been captured). Finally, we consider a discount factor $\gamma=0.95$.

\subsection{Pursuit under partial observability}

Our second evaluation scenario---the Pursuit domain under partial observability---is a modification of the original Pursuit domain where we add partial state observability. In it, $N$ agents ($N=2$ in our experiments) must capture a single moving prey by surrounding it in a coordinated way. In this scenario, the different tasks correspond to different behaviors of the teammate.

In this scenario, the ad hoc agent is able to observe the elements located in a larger neighborhood, but the probability of it failing to observe an element increases with the distance. Specifically, the agent fails to observe an existing element (teammate, prey) located at a distance $d$ with probability $\eps(d)=1-0.15 d$.

We model each possible task $m_k\in\M$ as a POMDP 
\begin{equation*}
(\X,\A^\alpha,\Z,\set{\P_{k,a^\alpha},a^\alpha\in\A^\alpha},\set{\O_{a^\alpha},a^\alpha\in\A^\alpha},r,\gamma).	
\end{equation*}
Each state $x\in\X$ contain information regarding the relative distances to the teammate and prey and is therefore represented by a tuple $x=(d{^{a_1}}_x, d{^{a_1}}_y, d{^p}_x, d{^p}_y)$, where $d{^{a_1}}_x, d{^{a_1}}_y$ represents the relative distance (in units) to the teammate and $d{^p}_x, d{^p}_y$ represents the relative distance (in units) to the prey. Each observation $z\in\Z$ is also represented as a tuple $z=(\hat{d}{^{a_1}}_x, \hat{d}{^{a_1}}_y, \hat{d}{^p}_x, \hat{d}{^p}_y)$, where the distance to an entity $e$ (prey or teammate), $(\hat{d}{^e}_x, \hat{d}{^e}_y)$, represents an observation of the respective true relative distance $(d{^e}_x, d{^e}_y)$. According to the probability $\eps(d)=1-0.15 d$, when a successful entity observation is made (i.e., roll $>= \eps(d)$), $(\hat{d}{^e}_x, \hat{d}{^e}_y) = (d{^e}_x, d{^e}_y)$. Otherwise, when an unsuccessful observation is made, (i.e., roll $< \eps(d)$), the tuple $(\hat{d}{^e}_x, \hat{d}{^e}_y)$ is instead filled with the distance to one of the four neighbouring cells to the entity (randomly picked). 
The action space $\A^\alpha$ contains four possible actions, \textit{Up, Down, Left, Right}. For each teammate type $k$, the transition probabilities $\set{\P_{k,a^\alpha},a^\alpha\in\A^\alpha}$ map a state $x$ and action $a$ to every possible next state $x'$, taking into account the probability of the teammate executing each possible action on $x$ given their policy for task $k$. Similarly, the observation probabilities $\set{\O_a^\alpha,a^\alpha\in\A}$ map a state $x$ and previous action $a^\alpha$ to every possible observation $z$, taking into account the probability $\epsilon(d)$ of the agent failing to observe the position of the other agents. The reward function $r_k$ assigns the reward of $-1$ for all time steps except those where the prey has been cornered, in which case it assigns a reward of $100$ for the one the prey was cornered in and $0$ afterwards (in other words, since where dealing with an horizon of 50 steps, we do not reset the environment whenever the prey have been cornered). Finally, we consider a discount factor $\gamma=0.95$.

\subsection{Overcooked}

Finally, for our third and last evaluation scenario, we resort to the Overcooked domain \citep{carroll19nips}. The Overcooked domain requires two agents, a helper and a cook, to cooperate in order to cook as many soups as possible in a given timeframe. In this case, tasks represent different teammates which the ad hoc agent must (i) identify; and (ii) adapt to, in order to cook as many soups as possible. Our ad hoc agent may play one of two possible roles: the {\em helper}, which has the goal of providing the cook with onions and plates, placing them on a kitchen counter; and the {\em cook}, which has the goal of cooking a soup using three onions and serving it in a plate (which is then dispatched through a window). The layout used in our experiments is depicted in Fig.~\ref{fig:overcooked}.

We can model each task $k$ in the Overcooked domain as an MDP $(\X,\A^\alpha,\{\P_{k,a^\alpha},a^\alpha\in\A^\alpha\},r,\gamma)$, since the state is fully observable in this domain. Each different task has distinct transition probabilities. Each state $x\in\X$ is represented as a tuple $x=(p_a, p_c, h_a, h_c, tb, bb, s)$, where, respectively, $(p_a, p_c)$ represents the cell (top or bottom) and $(h_a, h_c)$ represents the objects (nothing, onion, plate or soup) in hand of the helper and the cook. $(tb, bb)$ represents the objects on the top kitchen counter balcony and bottom kitchen counter balcony (respectively). Finally, $s$ represents the contents of the soup pan (empty, one onion, two onions, cooked soup). The action space $\A$ contains four possible actions, \textit{Up, Down, Noop} and \textit{Act}. The transition probabilities $\set{\P_{k,a^\alpha},a^\alpha\in\A^\alpha}$ map a state $x$ and action $a^\alpha$ to every possible next state $x'$, taking into account the probability of the teammate executing each action on $x$. The reward function $r$ assigns the reward of $15$ to states $x$ where the cook delivers a cooked soup through the kitchen window, and $-1$ otherwise. We consider a discount factor $\gamma=0.95$.


\end{document}